\documentclass{article}

\usepackage{microtype}
\usepackage{graphicx}
\usepackage{subfigure}
\usepackage{booktabs} 

\usepackage{algorithm}
\usepackage{algorithmicx}
\usepackage{algpseudocode}
\usepackage{thmtools,thm-restate}
\usepackage{comment}
\usepackage{mathtools}

\usepackage{hyperref}

\usepackage[accepted]{icml2019}

\usepackage{thmtools,thm-restate}
\usepackage{comment}

\usepackage{amsmath,amsfonts,bm}

\def\eqref#1{equation~\ref{#1}}

\def\1{\bm{1}}

\DeclareMathAlphabet{\mathsfit}{\encodingdefault}{\sfdefault}{m}{sl}
\SetMathAlphabet{\mathsfit}{bold}{\encodingdefault}{\sfdefault}{bx}{n}

\newcommand{\E}{\mathbb{E}}

\newcommand{\KL}{D_{\mathrm{KL}}}

\DeclareMathOperator*{\argmax}{arg\,max}

\newcommand{\vect}[1]{\mathbf{#1}}

\newcommand{\definedas}{\equiv}

\DeclareMathOperator{\renyiop}{D}
\newcommand{\renyi}[3]{\renyiop_{#1}\left(#2 \middle \| #3 \right)}

\newcommand{\sfr}{{\bm \phi}}
\newcommand{\sfQ}{{\bm \psi}}
\newcommand{\sfV}{{\bm \Upsilon}}

\usepackage{hyperref}
\usepackage{url}
\usepackage{hyperref}
\usepackage{graphicx}

\icmltitlerunning{Composing Entropic Policies using Divergence Correction}

\begin{document}

\twocolumn[
\icmltitle{Composing Entropic Policies using Divergence Correction}



\icmlsetsymbol{equal}{*}

\begin{icmlauthorlist}
\icmlauthor{Jonathan J Hunt}{dm}
\icmlauthor{Andre Barreto}{dm}
\icmlauthor{Timothy P Lillicrap}{dm}
\icmlauthor{Nicolas Heess}{dm}
\end{icmlauthorlist}

\icmlaffiliation{dm}{DeepMind}

\icmlcorrespondingauthor{Jonathan J Hunt}{jjhunt@google.com}

\icmlkeywords{Machine Learning, ICML}

\vskip 0.3in
]



\printAffiliationsAndNotice{}  

\begin{abstract}
%
Composing previously mastered skills to solve novel tasks promises dramatic improvements in the data efficiency of reinforcement learning. Here, we analyze two recent works composing behaviors represented in the form of action-value functions and show that they perform poorly in some situations. As part of this analysis, we extend an important generalization of policy improvement to the maximum entropy framework and introduce an algorithm for the practical implementation of successor features in continuous action spaces. Then we propose a novel approach which addresses the failure cases of prior work and, in principle, recovers the optimal policy during transfer. This method works by explicitly learning the (discounted, future) divergence between base policies. We study this approach in the  tabular case and on non-trivial continuous control problems with compositional structure and show that it outperforms or matches existing methods across all tasks considered.
%
\end{abstract}

\section{Introduction}

Reinforcement learning (RL) algorithms coupled with powerful function approximators have recently achieved a series of successes \citep{mnih2015human,silver2016mastering,lillicrap2015continuous, kalashnikov2018qt}.
Unfortunately, all of these approaches still require a large number of interactions with the environment. One reason for this is that the algorithms are typically applied ``from scratch,'' rather than leveraging experience from prior tasks.
This reduces their applicability in domains where generating experience is expensive, or learning from scratch is  challenging.


In humans, the development of complex motor skills, such as bipedal locomotion or high-speed manipulation, also requires large amounts of experience and practice
\citep{adolph2012you, haith2013model}
However, once such skills have been acquired humans rapidly put them to work in new contexts and to solve new tasks, 
suggesting transfer learning as an important mechanism for data efficiency.

Motivated by this observation, we are interested in RL methods for transfer that are suitable for high-dimensional motor control. We focus on model-free approaches which are evident in human motor control \citep{haith2013model} and underlie the recent successes of Deep RL cited above.


Transfer may be especially valuable in domains where a small set of skills can be composed, in different combinations, to solve a variety of tasks. Different notions of compositionality have been considered in the RL and robotics literature. For instance, `options' tend to be associated with discrete units of behavior that can be sequenced, thus emphasizing composition in time \citep{precup1998theoretical}. In this paper we are concerned with a rather distinct notion of compositionality, namely how to combine and blend potentially concurrent behaviors. This form of composition is particularly relevant in high-dimensional continuous action spaces, where it is possible to achieve more than one task simultaneously (e.g. walking somewhere while juggling).

One approach to this challenge is via the composition of task rewards. Specifically, we are interested in the following question: If we have previously solved a set of tasks with similar transition dynamics but different reward functions, how can we leverage this knowledge to solve new tasks which can be expressed as a convex combination of those reward functions?


This question has recently been studied in two independent lines of work: by \citet{barreto2017successor,barreto2018transfer} in the context of successor feature (SF) representations used for Generalized Policy Improvement (GPI) with deterministic policies, and by \citet{haarnoja2018composable,van2018will} in the context of maximum entropy (max-ent) policies. These approaches operate in distinct frameworks but both achieve skill composition by combining the $Q$-functions associated with previously learned skills.

In this paper, we clarify the relationship between the two approaches and show that both perform well in some situations but that they fail in complementary ways. We introduce a novel method of behavior composition that can consistently achieve good performance.

Our contributions are as follows:
\begin{enumerate}
    \item{We introduce succcessor features (SF) in the context of the maximum entropy framework and extend the GPI theorem to the max-ent objective (max-ent GPI).}

    \item We provide an analysis of when GPI, and compositional ``optimism'' (CO) \citep{haarnoja2018composable} perform poorly. We highlight these failure cases with tabular discrete action tasks and challenging continuous control tasks.

    \item We propose a correction term -- which we call Divergence Correction (DC)-- based on the R\'enyi divergence between policies which allows us, in principle, to recover the optimal policy for transfer for any convex combination of rewards.

    \item{We demonstrate a practical algorithm, which relies of adaptive importance sampling, for zero-shot transfer with DC or max-ent GPI in continuous action spaces. We compare the approaches introduced here, max-ent GPI and DC, with compositional optimism \citep{haarnoja2018composable} and Conditional $Q$ functions \citep{schaul2015universal} in a variety of non-trivial continuous action transfer tasks.}
\end{enumerate}

\section{Background}

\subsection{Multi-task RL}

We consider Markov Decision Processes defined by the tuple $\mathcal{M}$ containing:
a state space $\mathcal{S}$, action space $\mathcal{A}$, start state distribution $p(s_1)$, transition function $p(s_{t+1}|s_t, a_t)$, 
discount $\gamma \in [0, 1)$ and reward function $r(s_t, a_t, s_{t+1})$.
We aim to find an optimal policy $\pi(a | s): \mathcal{S} \rightarrow \mathcal{P}(\mathcal{A})$,
which is one that maximises the discounted expected return from any state $J(\pi) = \E_{\pi, \mathcal{M}}\left[ \sum_{\tau=t}^\infty \gamma^{\tau - t} r_\tau \right]$ where the expected return is dependent on the policy $\pi$ and the MDP $\mathcal{M}$.

We formalize transfer as in \citet{barreto2017successor, haarnoja2018composable}, as the desire to perform well across all tasks in a set $\mathcal{M} \in \mathcal{T}'$ (without direct experience on these tasks) after having learned policies for tasks $\mathcal{M} \in \mathcal{T}$.
We assume that $\mathcal{T}$ and $\mathcal{T}'$ are related in two ways: all tasks share the same state transition function, and tasks in $\mathcal{T}'$ can be expressed as convex combinations of rewards associated with tasks in set $\mathcal{T}$. We write the reward functions for tasks in $\mathcal{T}$ as the vector $\sfr = (r_1, r_2, \dots)$, so tasks in $\mathcal{T}'$ can be expressed as $r_\vect{w} = \sfr \cdot \vect{w}$. Our theoretical results assume that we have learned optimal policies for all tasks in $\mathcal{T}$.

For clarity, we focus on the combination of only two policies, that is $\vect{w}$ has only 2 non-zero entries and so we can write $r_b = b r_i + (1 - b) r_j$ ($b \in [0, 1]$). As we discuss later, the approaches we consider can be extended to more than two tasks. We refer to a transfer method as optimal, if it results in an optimal policy for the transfer task in $\mathcal{T}'$, using only experience on tasks $\mathcal{T}$. As in most prior work in Deep RL, theoretical guarantees of optimality can only be approximately achieved in practise.

\subsection{Successor Features}

Successor Features (SF) \citep{dayan1993improving} and Generalised Policy Improvement (GPI) \citep{barreto2017successor, barreto2018transfer} provide a principled solution to transfer in the setting defined above.
SF make the additional assumption that the reward feature $\sfr$ is fully observable, that is, the agent always has access to the rewards of all tasks in $\mathcal{T}$ but not $\mathcal{T}'$ during training.

The key insight of SF is that linearity of the reward $r_{\vect{w}}$ with respect to the features $\sfr$ implies the following decomposition of the action value of policy $\pi$ on task $r_\vect{w}$:
\begin{align*} \textstyle
    Q^{\pi}_\vect{w}(s_t, a_t) & 
     = \E_{\pi} \left[ \sum_{\tau=t}^\infty \gamma^{\tau - t} \sfr_\tau \cdot \vect{w} | a_t \right]
     \nonumber
     \\
    &= \E_{\pi} \left[ \sum_{i=t}^\infty \gamma^{\tau - t} \sfr_\tau | a_t \right] \cdot \vect{w} \definedas \sfQ^{\pi}(s_t, a_t) \cdot \vect{w}
\end{align*}
where $\sfQ^{\pi}$ is the expected discounted sum of features $\sfr$ induced by policy $\pi$. The SF decomposition allows us to compute the value of an existing policy $\pi$ on a new task $r_\vect{w}$.

GPI provides a principled way to use this value information to compose $n$ existing polices $\pi_1, \pi_2, ..., \pi_n$ indexed by $i$ to solve task $r_{\vect{w}}$.  Namely, we act according to the deterministic GPI policy $\pi^{GPI}_{\vect{w}}(s_t) \definedas \argmax_{a_t} Q^{GPI}_{\vect{w}}(s_t, a_t)$ where
\begin{align*} \textstyle
    Q^{GPI}_{\vect{w}}(s_t, a_t) \definedas \max_{i} Q^{\pi_i}_{\vect{w}}(s_t, a_t) = \max_i \sfQ^{\pi_i}(s, a) \cdot \vect{w}
\end{align*}
The GPI theorem guarantees the GPI policy has a return at least as good as any component policy, that is, $V^{\pi^{GPI}}_{\vect{w}}(s) \geq \max_i V^{\pi_i}_{\vect{w}}(s) \: \forall s \in \mathcal{S}$.

Note that SF and GPI are separate concepts, the GPI theorem does not require the use of SFs. SFs provide an efficient mechanism for computing the value of existing policies on a new task, and GPI provides a principled way to make use of this information to compose the existing policies.

\subsection{Maximum Entropy RL}

The maximum entropy (max-ent) RL objective augments the reward to favor entropic solutions
\begin{align} \textstyle
    J(\pi) =  \E_{\pi, \mathcal{M}}\left[ \sum_{\tau=t}^\infty \gamma^{\tau - t} (r_\tau + \alpha H[\pi(\cdot | s_\tau))] \right]
    \label{eq:maxent_objective}
\end{align}
where $\alpha$ is a parameter that determines the relative importance of the entropy term.

This objective has been considered in a number of works including \cite{kappen2005path,todorov2009compositionality, haarnoja2017reinforcement, haarnoja2018composable,ziebart2008maximum,fox2015taming}.

We define the action-value $Q^\pi$ associated with eq.\ \ref{eq:maxent_objective} as
\begin{align} \textstyle
\label{eq:soft_q}
    Q^\pi(s_t, a_t) \definedas r_t + \E_{\pi}\left[ \sum_{\tau={t+1}}^\infty \gamma^{\tau-t} (r_\tau + \alpha H[\pi(\cdot|s_\tau)])  \right]
\end{align}
(notice $Q^\pi(s_t, a_t)$ does not include the entropy term for the state $s_t$).
 Soft Q iteration
where the policy $\pi(a_t|s_t) \propto \exp (\frac{1}{\alpha} Q(s_t, a_t))$ is implicitly defined by $Q$
converges to the optimal policy with standard assumptions \citep{haarnoja2017reinforcement}.
\begin{align*} \textstyle
    Q(s_t, a_t) &\leftarrow r(s_t, a_t, s_{t+1}) + \gamma \E_{p(s_{t+1}|s_t, a_t)}\left[V(s_{t+1}) \right]
    \\
    V(s_t) &\leftarrow  \E_{{\pi}}\left[ Q(s_t, a_t) \right] + \alpha H[\pi(\cdot|s_t)]
    \\
    &= \alpha \log \int_{\mathcal{A}} \exp(\frac{1}{\alpha}Q(s_t, a_t))da \definedas \alpha \log Z(s_t) 
\end{align*}

\subsection{Compositional Optimisim (CO)}
\citet{haarnoja2018composable} introduced a simple approach to policy composition in the max-ent framework by approximating the optimal action-value for the transfer task $r_b = b r_i + (1 - b)r_j$ from the optimal action-values of the component tasks $Q^i$ and $Q^j$
\begin{align} \textstyle \label{eq:qopt}
    Q_b^{CO}(s, a) \definedas b Q^{i}(s, a) + (1 - b) Q^{j}(s, a).
\end{align}
When using Boltzmann policies defined by $Q$, the resulting policy,  $\pi_b^{CO}(a|s) \propto \exp(\frac{1}{\alpha}Q_b^{CO}(s, a))$, is the product distribution of the two component policies. We refer to $\pi^{CO}_b$ as the compositionally ``optimistic'' (CO) policy, as it acts according to an over-estimate of the action value ($Q_b^{CO}(s,a) \geq Q_b^*(s,a)$) by assuming the optimal returns of $Q^i$ and $Q^j$ will be, simultaneously, achievable.

\section{Composing Policies in Max-Ent Reinforcement Learning}

In this section we present two novel
approaches for transfer in the max-ent framework. In section \ref{sec:AIS} we then outline a practical algorithm using these results.

\subsection{Max-Ent Successor Features and Generalized Policy Improvement}

We introduce max-ent SF, which provide a mechanism for computing the value of a maximum entropy policy under any convex combination of rewards. We then show that the GPI theorem \citep{barreto2017successor} holds
for maximum entropy policies.

We define the action-dependent SF to include the entropy of the policy, excluding the current state, analogous to the max-entropy definition of $Q^\pi$ in~(\ref{eq:soft_q}):
\begin{align*} \textstyle
    \sfQ^\pi(s_t, a_t) &\definedas \sfr_t + \E_{\pi}\left[ \sum_{\tau={i+1}}^\infty \gamma^{\tau-t} (\sfr_\tau + \alpha \bm{1} \cdot H[\pi(\cdot|s)])\right] \\
    &= \phi_t + \gamma \E_{p(s_{t+1}|s_t, a_t)}\left[ \sfV(s_{t+1})\right]
\end{align*}
where $\bm{1}$ is a vector of ones of the same dimensionality as $\phi$ and we define the state-dependent successor features as the expected $\sfQ^\pi$ in analogy with $V^\pi(s)$:
\begin{align} \label{eq:sfv_maxent} \textstyle
    \sfV^\pi(s) \definedas \E_{a \sim \pi(\cdot |s)}\left[ \sfQ^\pi(s, a) \right] + \alpha \bm{1} \cdot H[\pi(\cdot|s)].
\end{align}
The max-entropy action-value of $\pi$ for any convex combination of rewards $\vect{w}$ is then given by $Q^\pi_{\vect{w}}(s, a) = \sfQ^\pi(s, a) \cdot \vect{w}$.
Max-ent SF allow us to estimate the action-value of previous policies on a new task. We show that, as in the deterministic case, there is a principled way to combine multiple policies using their action-values on task $\vect{w}$.

\begin{restatable}[Max-Ent Generalized Policy Improvement]{theorem}{maxentgpi}
Let $\pi_1, \pi_2, ..., \pi_n$ be $n$ policies with $\alpha$-max-ent action-value functions $Q^1, Q^2, ..., Q^n$ and value functions $V^1, V^2, ..., V^n$.
Define
\begin{align*} \textstyle
    \pi(a|s) \propto \exp\left(\frac{1}{\alpha} \max_i Q^i(s, a)\right).
\end{align*}
Then,
\begin{align} \textstyle
 \label{eq:gpiQ}
Q^\pi(s, a) &\geq \max_i Q^i(s, a) \text{ for all $s \in \mathcal{S}$, $a \in \mathcal{A}$,}
\\
V^\pi(s) &\geq \max_i V^i(s) \text{ for all $s \in \mathcal{S}$, }
 \label{eq:gpiV}
 \end{align}
where $Q^\pi(s, a)$ and $V^\pi(s)$ are the $\alpha$-max-ent action-value and value function respectively of $\pi$.
\end{restatable}
Proof: See appendix \ref{appendix:maxentgpi}.

In our setup, we learn $\sfQ^{\pi_i}(s,a)$, the SFs of policies $\pi_i$ for each task in $\mathcal{T}$, we define the max-ent GPI policy for task $\vect{w} \in \mathcal{T'}$ as
    %
 $   \pi_{\vect{w}}^{GPI}(a|s) \propto \exp(\frac{1}{\alpha} \max_i Q^{\pi_i}_{\vect{w}}(s, a) )
 = \exp(\frac{1}{\alpha} \max_i \sfQ^{\pi_i}(s, a) \cdot \vect{w}  ).$
In contrast to CO, GPI can be seen as acting ``pessimistically'' as it always acts according to a lower bound (\eqref{eq:gpiQ}) on the action-value.


\subsection{Divergence Correction (DC)} \label{section:ltd}

\begin{figure*}[t]
    \includegraphics[width=\textwidth]{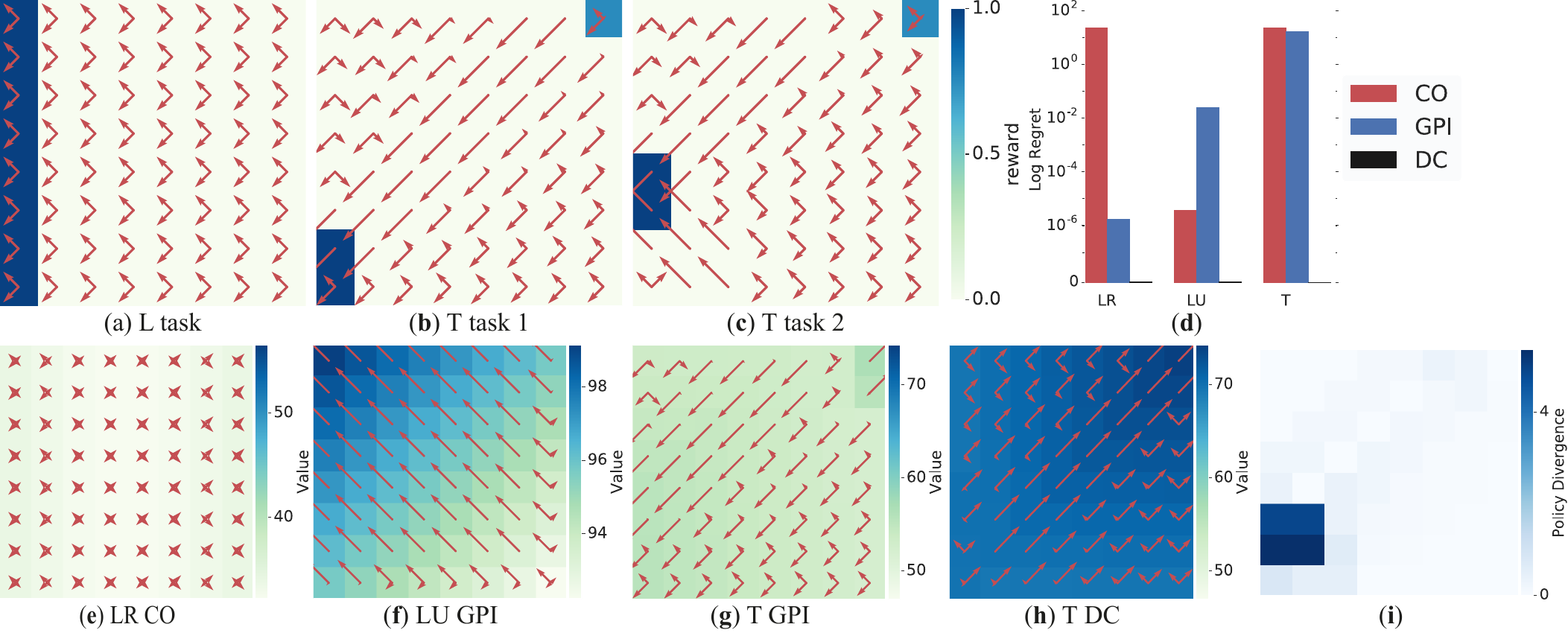}
    %
    \caption{\textbf{Policy composition in the tabular case}. All tasks are in an infinite-horizon tabular 8x8 world. The action space is the 4 diagonal movements (actions at the boundary transition back to the same state) (\textbf{a}-\textbf{c}) shows the reward functions for the L(eft) task and the two T(ricky) tasks (color indicates reward, dark blue $r=+1$, light blue $r=0.75$). The arrows indicate the action likelihoods for the max-ent optimal policy for each task.
     (\textbf{d}) The log regret on the compositional task $r_b = 1/2 r_i + 1/2r_j$ using the different methods for transfer. This is shown for three, qualitatively distinct compositional tasks: left-right (LR), left-up (LU) and the ``tricky`` tasks (T). GPI performs well on LR, where the subtasks are incompatible, meaning the optimal policy on the transfer task is similar to one of the existing policies. In LR CO fails to commit to a particular direction (\textbf{e} shows the CO policy and value) and performs poorly.
    Conversely, on the LU task when the base policies are compatible, CO transfers well while the GPI policy (\textbf{f}) does not consistently take advantage of the compatibility of the two tasks to simultaneously achieve both base rewards (unlike the CO policy suppl.\ figure \ref{fig:tabular_supp}c).
    Neither GPI nor CO policies (\textbf{g} shows the GPI policy, but CO is similar) perform well when the optimal transfer policy is unlike either existing task policy. The two tricky task policies are compatible in many states but have a high-divergence in the bottom-left corner since the rewards are non-overlapping there (\textbf{i} shows the divergence in each state of the two tricky base policies), thus the optimal policy on the compositional transfer task is to move to the top right corner where there are overlapping rewards. By learning, and correcting for, this future divergence between policies, DC results in optimal policies for all task combinations including tricky (\textbf{h}).
    Additional details in suppl.\ figure \ref{fig:tabular_supp}.
    }
    \label{fig:tabular}
\end{figure*}

Both max-ent GPI we presented above, and CO can, in different ways, fail to transfer well in some situations (fig.\  \ref{fig:tabular}). Neither approach consistently acts optimally during transfer, even if all component terms are known exactly.

Here we show, at the cost of learning a function conditional on the task weightings $b$, it is in principle possible to recover the optimal max-ent policy for the transfer tasks, without direct experience on those tasks, by correcting for the compositional optimism bias in $Q_b^{CO}$.

The correction term for CO uses a property noted, but not exploited in \citet{haarnoja2018composable}. The bias in $Q^{CO}$ is related to the the discounted sum of R\'enyi divergences of the two component policies. Intuitively, if the two policies induce trajectories with low divergence between the policies in each state, the CO assumption that both policies can achieve good returns simultaneously is approximately correct. When the divergences are large (i.e.\ the two policies do not agree on what action to take), the CO assumption is being overly optimistic.

\begin{restatable}[DC Optimality]{theorem}{ltdtheorem} \label{theorem:ltd}
    Let $\pi_i, \pi_j$ be $\alpha$ max-ent optimal policies for tasks with rewards $r_i$ and $r_j$ with max-ent action-value functions $Q^i, Q^j$.
    Define $C^\infty_b(s_t, a_t)$ as the fixed point of
    \begin{equation*}
    \begin{multlined}[][234.8775pt]
        {C^{(k+1)}_b(s_t, a_t) = -\alpha \gamma \E_{p(s_{t+1}|s_t, a_t)} \Big[  } \\
    {
    \log \int_\mathcal{A} \pi_i(a_{t+1}|s_{t+1})^b \pi_j(a_{t+1}|s_{t+1})^{(1 - b)} \exp \big(
    }
        \\
    {-\frac{1}{\alpha} C^{(k)}_b(s_{t+1}, a_{t+1})\big) da_{t+1} \Big]}
    \end{multlined}
    \end{equation*}
    %
    Given the conditions for Soft Q convergence,
the max-ent optimal $Q^*_b(s, a)$ for $r_b = b r_i + (1-b) r_j$ is
    \begin{align*} 
        Q^*_b(s, a) = b Q^i(s, a) + (1 - b) Q^j(s, a) - C^\infty_b(s, a)
        \\
        \forall s \in \mathcal{S}, a \in \mathcal{A}, b \in [0, 1]
    \end{align*}
\end{restatable}
Proof: See appendix \ref{appendix:ltd_proof}.

We call this Divergence Correction (DC) as the quantity $C^\infty_b$ is related to the R\'enyi divergence between policies (see appendix  \ref{appendix:ltd_proof} for details).
Learning $C^\infty_b$ does not require any additional information (in principle) than that required to learn policies $\pi_i$ and $\pi_j$. Unlike with SF, it is not necessary to observe other task rewards while training the policies. On the other hand, SF/GPI can be used to combine any number of tasks with arbitrary weight vectors $\vect{w}$, the difficulty of estimating $C^\infty$ increases significantly if more than two tasks are combined (see supplementary theorem A1).

Supplementary Table \ref{tab:comparison} provides a comparison on the properties of the methods we consider here. We also compare with simply learning a conditional $Q$ function $Q(s, a | b)$ (CondQ) \citep[e.g.][]{schaul2015universal, andrychowicz2017hindsight}. As with GPI, this requires observing the full set of task features $\sfr$, in order to compute $r_b$ for arbitrary $b$.

We have introduced two new approaches to max-ent transfer composition and described their properties: max-ent SF/GPI and DC. Now we address the question of how to practically learn and sample with these approaches in continuous action spaces.

\section{Adaptive Importance Sampling for Boltzman Policies Algorithm}
\label{sec:AIS}


Robotic systems with high-dimensional continuous action spaces are promising use cases for the ideas presented above, particularly as data efficiency is often a paramount concern. Such control problems may allow for multiple solutions, and often contain exploitable compositional structure.

Unfortunately, learning and sampling of general Boltzmann policies defined over continuous action spaces is challenging.
One approach is to fit an expressible, tractable sampler, such as a stochastic neural network to approximate $\pi_i$ \citep[e.g.][]{haarnoja2018composable}. This approach works well when learning a single policy. However, during transfer this may require learning a new sampler for each new transfer task.
Here we want to zero-shot transfer by sampling from a newly synthesized action-value function online, which precludes fitting a sampler. To address this issue we introduce Adaptive Importance Sampling for Boltzmann Policies (AISBP), which provides a practical solution to this challenge.


We use parametric approximators (e.g.\ neural networks) and denote their parameters $\theta$, including the soft action-value for reward $i$: $Q^i_{\theta_Q}(s, a)$; the associated soft value function $V^i_{\theta_V}(s)$ and a proposal distribution $q^i_{\theta_q}(a|s)$, which is used for importance sampling the policy (we will sometimes drop the task index $i$ for notational simplicity, and write the losses for a single policy).

We use an off-policy algorithm, so that experience generated by training on policy $\pi_i$ can be used to improve policy $\pi_j$. This is important to ensure that $Q^j_{\theta_Q}(s,a)$ is a good approximation of the action-value function in states that are likely under $\pi_j$ but unlikely under $\pi_i$ (suppl.\ \ref{appendix:experiment_details} discusses issues of exploration and coverage of the state space under function approximation).
Training experience across all tasks is stored in a replay buffer $R$, and mini-batches of experience are sampled uniformly from the replay.

The proposal distribution is a mixture of $M$ truncated Normal distributions $\mathcal{N}_T$, truncated to the square $a \in [-1, 1)^n$ with diagonal covariances.
\begin{align} \textstyle \label{eq:mixture_of_normal}
    q_{\theta_q}(a|s) = \frac{1}{M}\sum_{m=1}^M \mathcal{N}_T(a; \mu^m_{\theta_q}(s), \sigma^m_{\theta_q}(s), -1, 1)
\end{align}

The proposal distribution is optimized by minimizing the forward KL divergence with the Boltzmann policy $\pi(a|s) \propto \exp \frac{1}{\alpha} Q_{\theta_Q}(s, a)$. This KL is ``zero avoiding'' and over-estimates the support of $\pi$ \citep{Murphy2012} which is desirable for a proposal distribution \citep{gu2015neural}. The proposal loss is 
\begin{align} \textstyle
    \mathcal{L}(\theta_q) &= \E_{R}\left[ \E_{a \sim \pi(\cdot|s)}[\log \pi(a|s_t) - \log q_{\theta_q}(a|s_t)] \right]
\label{eq:loss_qq}
\end{align}
where the expectation is over the replay buffer state density.

The inner expectation in the proposal loss itself requires sampling from $\pi$. We approximate this expectation by self-normalized importance sampling and
use a target proposal distribution 
$p(a_t | s_t)$
which is a mixture distribution consisting of the proposals for all policies along with a uniform distribution. 
%
For batchsize $B$ and $N$ proposal action samples the estimator of the proposal loss is then
\begin{equation}
\begin{split} \textstyle
    \mathcal{L}(\theta_q) & \approx -\frac{1}{B }\sum_{k=1}^{B} \sum_{l=1}^N w_{kl} \log q_{\theta_q}(a_{kl}|s_k)
    \nonumber
    \\
    w'_{kl} &= \frac{\exp \frac{1}{\alpha}(Q_{\theta_Q}(s_k, a_{kl})) }{p(a_{kl}| s_k)} ; \: \: w_{kl} = \frac{w'_{kl}}{\sum_{m=1}^N w'_{km}} \label{eq:proposal_update2}
\end{split}
\end{equation}

By restricting the proposal distribution to mixtures of (truncated) Gaussians, we ensure the product of the proposal distributions is tractable. We make use of this product proposal during transfer (see supplementary \ref{subsec:transfer_sampling}).



%
%


%
The policy is improved using Soft Q iteration (eq.\ \ref{eq:soft_q}).
The value function loss is defined as the L2 error on the Soft Q estimate of value
\begin{equation}
\begin{split} \textstyle
    \mathcal{L}(\theta_V) =
     \: \E_{R}\Big[ \frac{1}{2}\Big(&V_{\theta_V}(s_t) - \\ & \alpha \log \int_{\mathcal{A}} \exp(\frac{1}{\alpha}Q_{\theta_Q}(s_t, a)) da \Big)^2 \Big] \label{eq:valueloss}
\end{split}
\end{equation}
which is estimated using importance sampling to compute the integral.
\begin{equation}
\begin{split} \textstyle
    \mathcal{L}(\theta_V) \approx \frac{1}{2B}\sum_{l=1}^B \left(V_{\theta_V}(s_l) - \alpha \log Z \right)^2
    \\
    Z = \left[\frac{1}{N} \sum_{k=1}^N \frac{\exp(\frac{1}{\alpha} Q_{\theta_Q}(s_{l}, a_{lk}))}{q_{\theta_q}(a_{lk}|s_l)} \right] \label{eq:alpha_logz_est}
\end{split}
\end{equation}
This introduces bias due to the finite-sample approximation of the expectation inside the (concave) $\log$. In practice we found this estimator sufficiently accurate, provided the proposal distribution was close to $\pi$. We also use importance sampling to sample from $\pi$ while acting.

The action-value loss 
is the L2 norm with the Soft Q 
target:
\begin{align} \textstyle
    \mathcal{L}(\theta_Q) = \: \E_{R}\Big[\frac{1}{2}(&Q_{\theta_Q}(s_t, a_t) -
    \notag \\
    &(r(s_t, a_t, s_{t+1}) + \gamma V_{\theta_V'}(s_{t+1})))^2 \Big]
\end{align}

To improve stability we employ target networks for the value $V_{\theta_{V'}}$ and proposal $q_{\theta_q'}$ networks \citep{mnih2015human, lillicrap2015continuous}
We also parameterize $Q$ as an advantage
$Q_{\theta_Q}(s, a) = V_{\theta_V}(s) + A_{\theta_A}(s, a)
$ \citep{baird1994reinforcement, wang2015dueling, harmon1995advantage} which is more stable when the advantage is small compared with the value. 
The full algorithm is given in Algorithm \ref{algo:basic} and more details are provided in appendix \ref{appendix:algo_details}.

\begin{algorithm}[h]
    \begin{algorithmic}
        \State Initialize proposal network parameters $\theta_q$
        \State Initial value network parameters $\theta_{V}$
        \State Initialize action-value network parameters $\theta_Q$
        \State Initialize replay $R$
        \While{training} \Comment{in parallel on each actor}
            \State{Obtain parameters $\theta$ from learner}
            \State{Sample task $i \sim \mathcal{T}$} \State{Importance sample $\pi_i(a|s) \propto \exp \frac{1}{\alpha} Q^i_{\theta_Q}(s, a)$}
            \State \: \: Using proposal $q^i_{\theta_q}$
            \State{Add experience to replay $R$}
        \EndWhile
        \While{training} \Comment{in parallel on the learner}
            \State Sample SARS tuple from $R$
            \State Improve $\mathcal{L}(\theta_q)$,
            $\mathcal{L}(\theta_V)$,
            $\mathcal{L}(\theta_Q)$
            \State Improve additional losses for transfer
            \State \: \: $\mathcal{L}(\theta_{\sfV})$, $\mathcal{L}(\theta_{\sfQ})$,  $\mathcal{L}(\theta_{C})$,
            $\mathcal{L}(\theta_{V_b})$
            $\mathcal{L}(\theta_{Q_b})$,
            \If{target update period}
                \State Update target network parameters \State \: \: $\theta_{V'} \leftarrow \theta_V, \theta_{q'} \leftarrow \theta_{q}$, $\theta_{\sfV'} \leftarrow \theta_{\sfV}$, $\theta_{V_b'} \leftarrow \theta_{V_b}$
            \EndIf
        \EndWhile
    \end{algorithmic}
    \caption{AISBP training algorithm \label{algo:basic}}
\end{algorithm}

\subsection{Importance Sampled Max-Ent GPI}

The same importance sampling approach can also be used to estimate max-ent SFs. Max-ent GPI requires us to learn the expected (maximum entropy) features $\sfQ_i$ for each policy $\pi_i$, in order to estimate its (entropic) value under a new convex combination task $\vect{w}$.
This requires that the experience tuples in the replay contain the full feature vector $\sfr$, rather than just the reward for the policy which generated the experience $r_i$. Given this information $\sfQ_{\theta_\sfQ}$ and $\sfV_{\theta_\sfV}$ can be learned with analogous updates to $V$ and $Q$. 

As with $V_{\theta_V}$, we use a target network for $\sfV_{\theta_\sfV'}$ and advantage parametrization.
We found it more stable to using a larger target update period than for $V$. Full details are of the losses and samplers are in appendix \ref{appendix:algo_details}.

\begin{figure}[t]
    \centering
    \includegraphics[width=\columnwidth]{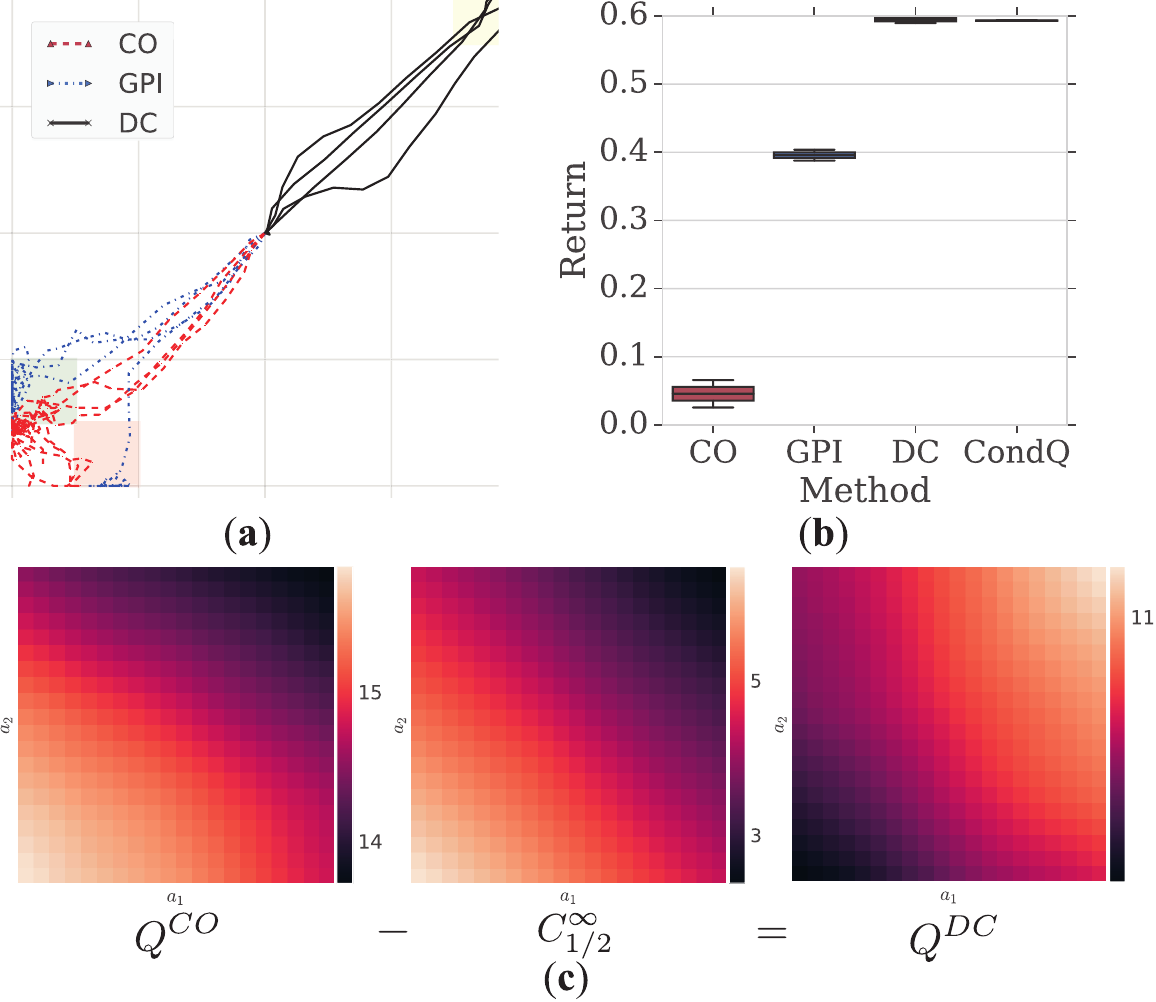}
    \caption{\textbf{Tricky point mass}. The continuous ``tricky'' task with a 2-D velocity controlled pointed mass.
    (\textbf{a}) Environment and example trajectories. The rewards are $(r_1=1, r_2=0)$, $(0, 1)$ and $(0.75, 0.75)$ for the green, red and yellow squares respectively. Lines show sampled trajectories (starting in the center) for the compositional task $r_{1/2}$ with different transfer methods. Only DC and CondQ (not shown) navigate to the yellow reward area during transfer, which is the optimal reward.
    (\textbf{b}) Box plot of returns for each transfer method (5 seeds). DC and CondQ methods perform significantly better than GPI, and the CO policy performs poorly.
    (\textbf{c}) $Q^{CO}$ at the center position for the transfer task. As both base policies prefer moving left and down, most of the energy (brighter color) is on these actions. However, the future divergence $C_{1/2}^\infty$ under these actions is high, which results in the $Q^{DC}$ differing qualitatively from CO and favoring the upward trajectory.
    Additional details in supplementary figure \ref{fig:point_mass_sup}.
}
    \label{fig:point_mass}
\end{figure}

\subsection{Divergence Correction}

Transfer using compositional optimism (\eqref{eq:qopt}, \citet{haarnoja2018composable}) only requires the max-ent action values of each task, so no additional training is required beyond the base policies.
In section \ref{section:ltd} we have shown that if we can learn the fixed point of $C^\infty_b(s, a)$ we can correct this compositional optimism bias and recover the optimal action-value $Q^*_b(s, a)$ for the transfer task $r_b$.

We exploit the recursive relationship in $C^\infty_b(s, a)$ to fit a neural net $C_{\theta_C}(s, a, b)$ with a TD(0) estimator. This requires learning a conditional estimator for any value of $b$, so as to support arbitrary task combinations. Fortunately, the same experience can be used to learn an estimator for all values of $b$, by sampling $b$ during each TD update.
We learn an estimator $C_{\theta_C}(s, a, b)$ for each pair of policies $\pi_i, \pi_j$ with the loss
\begin{equation}
\begin{split} \textstyle
    \mathcal{L}(\theta_C) =& \E_{s \sim R, b \sim U(0, 1)} [ \frac{1}{2} (
    C_{\theta_C}(s, a, b) +
    \\
    &\alpha \gamma \E_{p(s'|s, a)}[ \log \int_{\mathcal{A}} \exp(b \log \pi_i(a'|s') + \\ &(1 - b) \pi_j(a'|s') - \frac{1}{\alpha}C_{\theta_{C'}}(s', a', b)) da' ]
    )^2 ]
\end{split}
\end{equation}
As with other integrals of the action space, we approximate this loss using importance sampling to estimate the integral. As before, we use target networks and an advantage parametrization for $C_{\theta_C}(s, a, b)$
%
Note that, unlike GPI and CondQ (next section), learning $C^{\infty}_b$ does not require observing $\phi$ while training.

We also considered a heuristic approach where we learned $C$ only for $b=\frac{1}{2}$ (this is typically approximately the largest divergence). This avoids the complexity of a conditional estimator and then we we estimate $C_b^\infty$ during transfer as
 $   \hat{C}_b^\infty(s, a) \sim 4 b (1 - b) C_{1/2}^\infty(s, a). $
This heuristic, we denote DC-Cheap, can be motivated by considering Gaussian policies (see appendix \ref{appendix:ltd_heuristic})
The max-ent GPI bound can be used to correct for over-estimates of the heuristic $C_b^\infty$,
$    Q^{DC-Cheap+GPI}(s, a) = \max(Q^{CO}(s, a) - \hat{C}^\infty_b(s, a), Q^{GPI}(s, a))
$.

\subsection{Cond Q}

As a baseline, we learn a conditional $Q$ function using a similar approach to DC of sampling $b$ each update $Q(s, a, b)$ \citep{schaul2015universal}. This, like GPI but unlike DC, requires observing $\phi$ during training so the reward on task $b$ can be estimated. Full details provided in appendix \ref{appendix:algo_details}.


\begin{figure}[t]
    \centering
    \includegraphics[width=0.86\columnwidth]{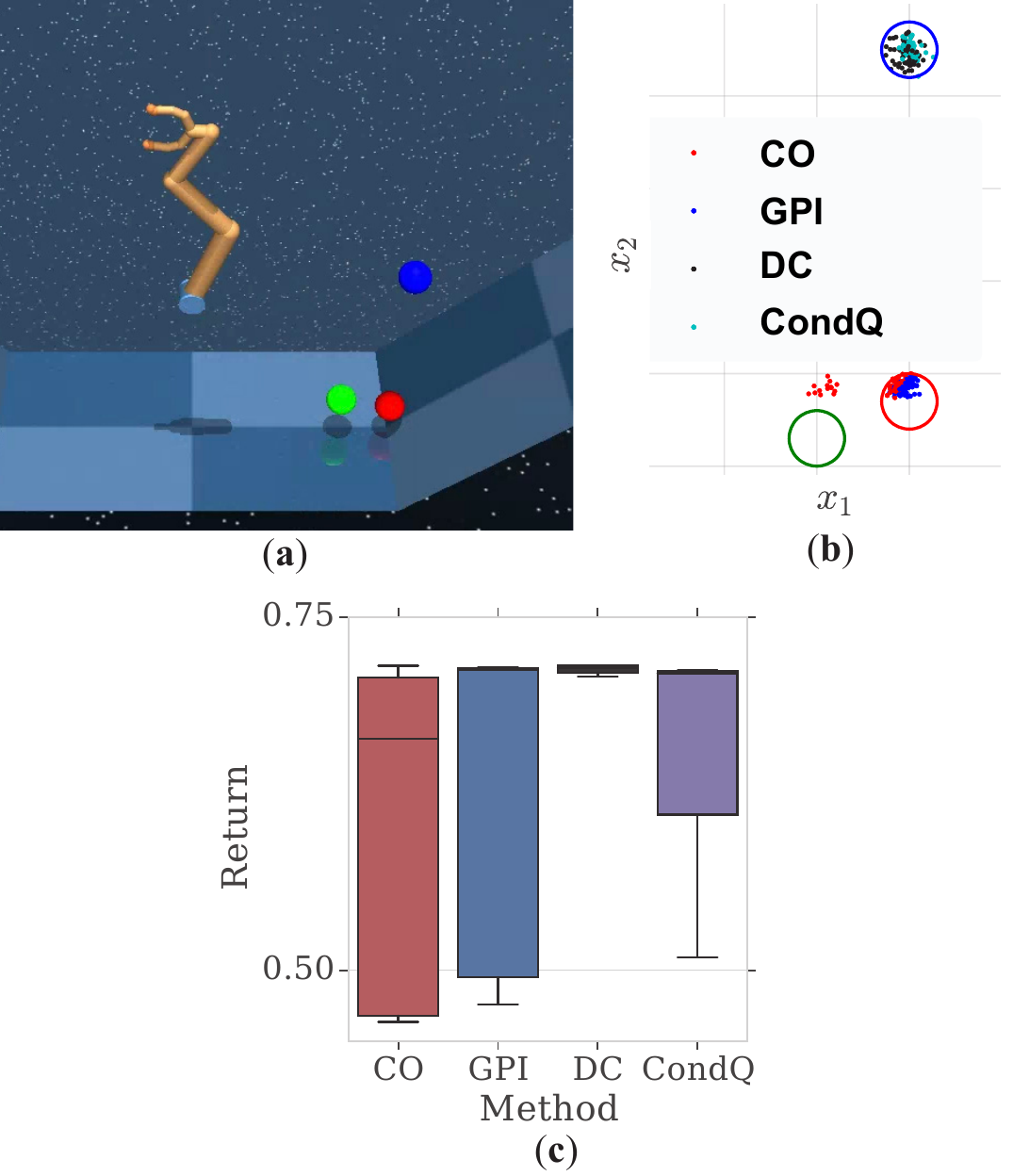}
    \caption{\textbf{``Tricky'' task with planar manipulator}. (\textbf{a}) The ``tricky'' tasks with a 5D torque-controlled planar manipulator. The training tasks consists of (mutually exclusive) rewards of $(1, 0), (0, 1)$ when the finger is at the green and red targets respectively and reward $(0.75, 0.75)$ at the blue target. (\textbf{b}) Finger position at the end of the trajectories starting from randomly sampled start states) for the transfer task with circles indicating the positions of the targets. DC and CondQ trajectories reach towards the blue target (the optimal solution) while CO and GPI trajectories primarily reach towards one of the suboptimal partial solutions. (\textbf{c}) Box plot of returns on the transfer tasks, DC outperforms other methods.
    Additional details in supplementary figure \ref{fig:planar_manipulator_supp}.
    }
    \label{fig:reacher}
\end{figure}

\subsection{Sampling Compositional Policies}
\label{sec:SamplingCompositional}

During transfer we need to sample from the Boltzmann policy defined by our estimate of the transfer action-value $Q_b$ (the estimate is computed using the methods we enumerated above). We wish to avoid needing to, offline, learn a new proposal or sampling distribution first (which  is the approach employed by \citet{haarnoja2018composable}).

As outlined earlier (and detailed in appendix \ref{subsec:transfer_sampling}), we chose the proposal distributions so that the product of proposals is tractable, meaning we can sample from $q^{ij}_b(a|s) \propto (q^i_{\theta_q}(a|s))^b (q^j_{\theta}(a|s))^{(1-b)}$. This is a good proposal distribution when the CO bias is low, since $Q^{CO}_b$ defines a Boltzmann policy which is the product of the base policies\footnote{$\pi^{CO}_b(a|s) \propto \exp \frac{1}{\alpha}Q^{CO}(s, a) = \exp (\frac{1}{\alpha}(Q^{1}(s, a) + Q^{2}(s,a)) = \pi_1(a|s) \pi_2(a|s)$.}.
However, when $C^\infty_b(s, a)$ is large, meaning the CO bias is large, $q^{ij}$ may not be a good proposal, as we show in the experiments. In this case none of the existing proposal distributions may be a good fit. Therefore we sample from a mixture distribution of all policies, all policy products and the uniform distribution, with the hope that at least one component of this mixture will be sampling from high-value parts of the action space.
\begin{align} \textstyle
    p_b(a|s) \definedas \frac{1}{4}(q_{\theta_q}^i(a|s) + q_{\theta_q}^j(a|s) + q_b^{ij}(a|s) + \frac{1}{\mathcal{V}^A})
\end{align}
where $\mathcal{V}^A$ is the volume of the action space. Empirically, we find this is sufficient to result in good performance during transfer. The transfer algorithm is given in supplementary algorithm \ref{algo:transfer}.

\section{Experiments}

\subsection{Discrete, tabular environment} \label{section:tabular}
We first provide some illustrative tabular cases of compositional transfer. These highlight situations in which GPI and CO transfer can perform poorly (Figure \ref{fig:tabular}). 
As expected from theory, we find that GPI performs well when the optimal transfer policy is close to one of the existing policies; CO performs well when both subtask policies are compatible, meaning there is some part of the action space that has a high likelihood under both policies.
%
The task we refer to as ``tricky'' is illustrative of challenging tasks we will focus on in the next section, namely scenarios
in which the optimal policy for the transfer task does not resemble either existing policy:
In the grid world non-overlapping rewards for each task are provided in one corner of the grid world, while lower value overlapping rewards are provided in the other corner (cf.~Fig.~\ref{fig:tabular}). As a consequence both GPI and CO perform poorly while DC performs well in all cases.

\begin{figure}[t]
    \centering
    \includegraphics[width=0.86\columnwidth]{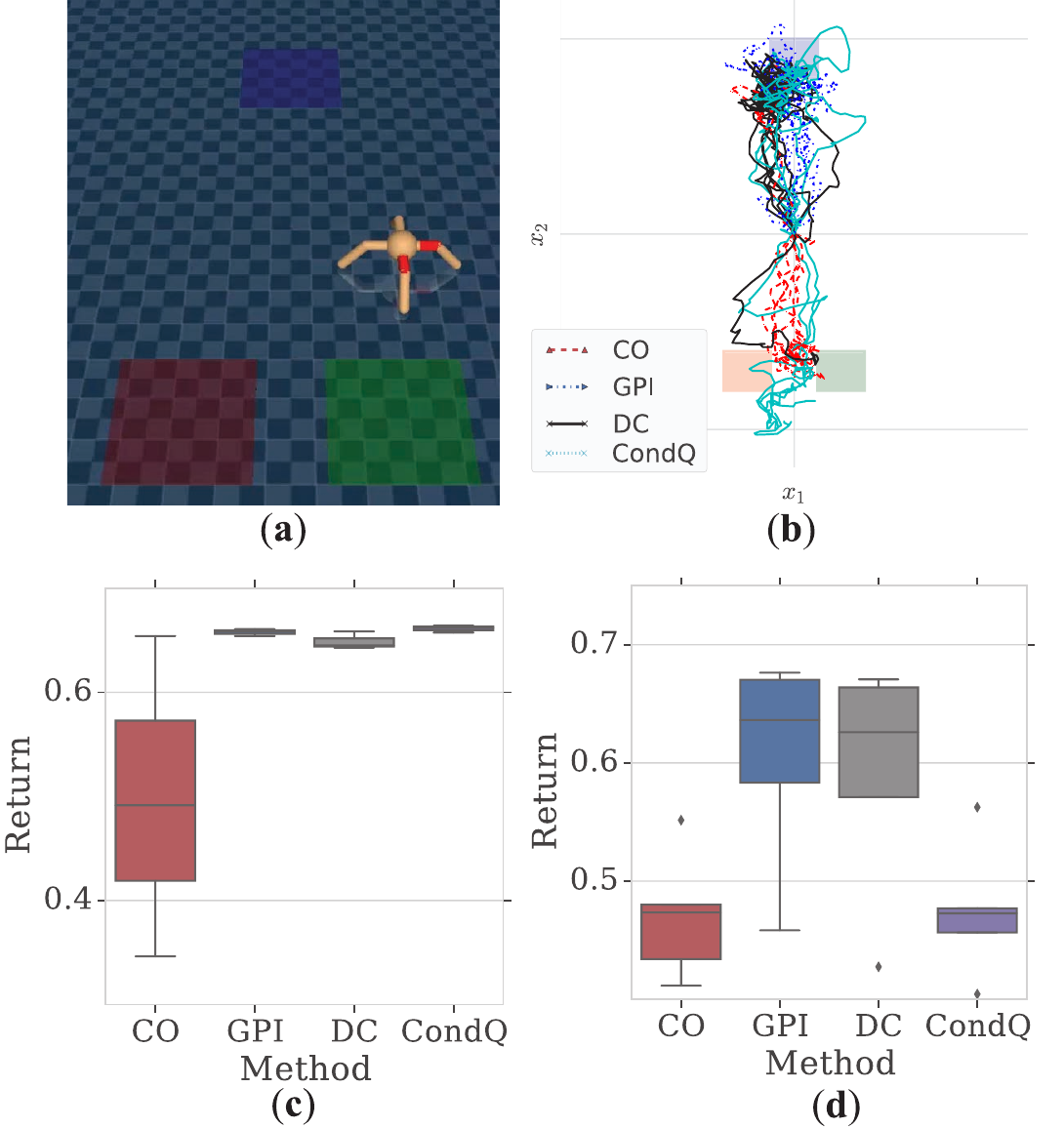}
    \caption{\textbf{``Tricky'' task with mobile bodies.} ``Tricky'' task with two bodies: a 3 DOF jumping ball (supplementary figure \ref{fig:walker_supp}a) and (\textbf{a}) 8 DOF ant (both torque controlled). The task has rewards $(1, 0), (0, 1)$ in the green and red boxes respectively and $(0.75, 0.75)$ in the blue square.
    (\textbf{b}) Sampled trajectories of the ant on the transfer task ($b=\frac{1}{2}$) starting from a neutral position. GPI and DC almost always go to the blue square (optimal), CondQ and CO do not.
    Box plot of returns for the jumping ball (\textbf{c}) and ant (\textbf{d}) when started in the center position. CO does not recover a good transfer policy for the compositional task while the other approaches largely succeed, although CondQ does not learn a good policy on the ant.
    Additional details in supplementary figure \ref{fig:walker_supp}.
    }
    \label{fig:jumping_ball_tricky}
\end{figure}

\subsection{Continuous action spaces}

In the tabular environments we demonstrated some challenging ``tricky'' transfer tasks that previous methods perform poorly on. Here, we test our approach on continuous control tasks with the same challenging properties. We train max-ent policies to solve individual tasks and then compare transfer performance using the different approaches. All approaches use the same experience, proposal distributions and base policies.


Figure \ref{fig:point_mass} examines the transfer policies in detail in a simple point-mass version of the ``tricky'' tasks and shows how the estimated $C^{\infty}_b$ corrects $Q^{CO}$ and results in a qualitatively better transfer policy.

We now examine conceptually similar tasks in more difficult domains: a 5 DOF planar manipulator reaching task (figure \ref{fig:reacher}), 3 DOF jumping ball and 8 DOF ant (figure \ref{fig:jumping_ball_tricky}). We see that DC recovers a qualitatively better policy in all cases. The performance of GPI depends noticeably on the choice of $\alpha$. DC-Cheap, which is a simpler heuristic, performs almost as well as DC in the tasks we consider except for the point mass task. When bounded by GPI (DC-Cheap+GPI) it performs well for the point mass task as well, suggesting these heuristics may be sufficient in some cases\footnote{
 Videos of the tasks and supplementary information at \url{https://tinyurl.com/yaplfwaq}.
}.

We focused on ``tricky'' tasks as they are a particularly challenging form of transfer. There are two other cases we considered in the tabular analysis. One: when the base tasks are compatible, CO performs well. We expect that DC would also perform well in this situation since, in this case, the correction term $C_b^{\infty}$ that DC must learn is inconsequential (CO is equivalent to assuming $C_b^{\infty}=0$). Two: At the other extreme, supplementary figure \ref{fig:supp_gpi} demonstrates on a task with incompatible base tasks (i.e. $C_b^{\infty}$ is large and potentially challenging to learn), DC continues to perform as well as GPI, slightly better than CondQ and much better than CO. In principle, without function approximator error, DC always recovers the optimal policy during transfer. These experiments provide empirical evidence that our approximate algorithm performs well in a range of situations.

\section{Discussion}

We have presented two approaches to transfer learning via convex combinations of rewards in the maximum entropy framework: max-ent GPI and DC. We have shown that, under standard assumptions, the max-ent GPI policy performs at least as well as its component policies, and that DC recovers the optimal transfer policy. \citet{todorov2009compositionality} and \citep{pan2015sample, saxe2017hierarchy, van2018will} previously considered optimal composition of max-ent policies. However, these approaches require stronger assumptions about the class of MDPs.
By contrast, DC does not restrict the class of MDPs and {\em learns} how compatible policies are, allowing approximate recovery of optimal transfer policies both when the component rewards are jointly achievable (AND), and when only one sub-goal can be achieved (OR).

We have compared our methods with conditional action-value functions (CondQ) \citep[e.g.]{schaul2015universal} and optimistic policy combination \citep{haarnoja2018composable}. Further, we have presented AISBP, a practical algorithm for training DC and max-ent GPI models in continuous action spaces using adaptive importance sampling. We have compared these approaches, along with heuristic approximations of DC, and demonstrated that DC recovers an approximately optimal policy during transfer across a variety of high-dimensional control tasks. Empirically we have found CondQ may be harder to learn than DC, and it requires additional observation of $\phi$ during training.

\newpage

\section*{Acknowledgements}

We thank Alexander Pritzel, Alistair Muldal,  Dhruva Tirumala, Siqi Liu and the rest of the DeepMind team for helpful discussions and assistance with this work.

\bibliography{icml_2019}

\begin{thebibliography}{32}
\providecommand{\natexlab}[1]{#1}
\providecommand{\url}[1]{\texttt{#1}}
\expandafter\ifx\csname urlstyle\endcsname\relax
  \providecommand{\doi}[1]{doi: #1}\else
  \providecommand{\doi}{doi: \begingroup \urlstyle{rm}\Url}\fi

\bibitem[Adolph et~al.(2012)Adolph, Cole, Komati, Garciaguirre, Badaly,
  Lingeman, Chan, and Sotsky]{adolph2012you}
Adolph, K.~E., Cole, W.~G., Komati, M., Garciaguirre, J.~S., Badaly, D.,
  Lingeman, J.~M., Chan, G.~L., and Sotsky, R.~B.
\newblock How do you learn to walk? thousands of steps and dozens of falls per
  day.
\newblock \emph{Psychological science}, 23\penalty0 (11):\penalty0 1387--1394,
  2012.

\bibitem[Andrychowicz et~al.(2017)Andrychowicz, Wolski, Ray, Schneider, Fong,
  Welinder, McGrew, Tobin, Abbeel, and Zaremba]{andrychowicz2017hindsight}
Andrychowicz, M., Wolski, F., Ray, A., Schneider, J., Fong, R., Welinder, P.,
  McGrew, B., Tobin, J., Abbeel, O.~P., and Zaremba, W.
\newblock Hindsight experience replay.
\newblock In \emph{Advances in Neural Information Processing Systems}, pp.\
  5048--5058, 2017.

\bibitem[Baird(1994)]{baird1994reinforcement}
Baird, L.~C.
\newblock Reinforcement learning in continuous time: Advantage updating.
\newblock In \emph{Neural Networks, 1994. IEEE World Congress on Computational
  Intelligence., 1994 IEEE International Conference on}, volume~4, pp.\
  2448--2453. IEEE, 1994.

\bibitem[Barreto et~al.(2017)Barreto, Dabney, Munos, Hunt, Schaul, van Hasselt,
  and Silver]{barreto2017successor}
Barreto, A., Dabney, W., Munos, R., Hunt, J.~J., Schaul, T., van Hasselt,
  H.~P., and Silver, D.
\newblock Successor features for transfer in reinforcement learning.
\newblock In \emph{Advances in neural information processing systems}, pp.\
  4055--4065, 2017.

\bibitem[Barreto et~al.(2018)Barreto, Borsa, Quan, Schaul, Silver, Hessel,
  Mankowitz, Zidek, and Munos]{barreto2018transfer}
Barreto, A., Borsa, D., Quan, J., Schaul, T., Silver, D., Hessel, M.,
  Mankowitz, D., Zidek, A., and Munos, R.
\newblock Transfer in deep reinforcement learning using successor features and
  generalised policy improvement.
\newblock In \emph{Proceedings of the International Conference on Machine
  Learning}, pp.\  501--510, 2018.

\bibitem[Clevert et~al.(2015)Clevert, Unterthiner, and
  Hochreiter]{clevert2015fast}
Clevert, D.-A., Unterthiner, T., and Hochreiter, S.
\newblock Fast and accurate deep network learning by exponential linear units
  (elus).
\newblock \emph{arXiv preprint arXiv:1511.07289}, 2015.

\bibitem[Dayan(1993)]{dayan1993improving}
Dayan, P.
\newblock Improving generalization for temporal difference learning: The
  successor representation.
\newblock \emph{Neural Computation}, 5\penalty0 (4):\penalty0 613--624, 1993.

\bibitem[Fox et~al.(2016)Fox, Pakman, and Tishby]{fox2015taming}
Fox, R., Pakman, A., and Tishby, N.
\newblock Taming the noise in reinforcement learning via soft updates.
\newblock In \emph{Proceedings of the Thirty-Second Conference on Uncertainty
  in Artificial Intelligence}, UAI'16, pp.\  202--211, Arlington, Virginia,
  United States, 2016. AUAI Press.
\newblock ISBN 978-0-9966431-1-5.
\newblock URL \url{http://dl.acm.org/citation.cfm?id=3020948.3020970}.

\bibitem[Gil et~al.(2013)Gil, Alajaji, and Linder]{gil2013renyi}
Gil, M., Alajaji, F., and Linder, T.
\newblock R{\'e}nyi divergence measures for commonly used univariate continuous
  distributions.
\newblock \emph{Information Sciences}, 249:\penalty0 124--131, 2013.

\bibitem[Gu et~al.(2015)Gu, Ghahramani, and Turner]{gu2015neural}
Gu, S., Ghahramani, Z., and Turner, R.~E.
\newblock Neural adaptive sequential monte carlo.
\newblock In \emph{Advances in Neural Information Processing Systems}, pp.\
  2629--2637, 2015.

\bibitem[Haarnoja et~al.(2017)Haarnoja, Tang, Abbeel, and
  Levine]{haarnoja2017reinforcement}
Haarnoja, T., Tang, H., Abbeel, P., and Levine, S.
\newblock Reinforcement learning with deep energy-based policies.
\newblock In \emph{International Conference on Machine Learning}, 2017.

\bibitem[Haarnoja et~al.(2018{\natexlab{a}})Haarnoja, Pong, Zhou, Dalal,
  Abbeel, and Levine]{haarnoja2018composable}
Haarnoja, T., Pong, V., Zhou, A., Dalal, M., Abbeel, P., and Levine, S.
\newblock Composable deep reinforcement learning for robotic manipulation.
\newblock In \emph{International Conference on Robotics and Automation},
  2018{\natexlab{a}}.

\bibitem[Haarnoja et~al.(2018{\natexlab{b}})Haarnoja, Zhou, Abbeel, and
  Levine]{haarnoja2018soft}
Haarnoja, T., Zhou, A., Abbeel, P., and Levine, S.
\newblock Soft actor-critic: Off-policy maximum entropy deep reinforcement
  learning with a stochastic actor.
\newblock In \emph{International Conference on Machine Learning},
  2018{\natexlab{b}}.

\bibitem[Haith \& Krakauer(2013)Haith and Krakauer]{haith2013model}
Haith, A.~M. and Krakauer, J.~W.
\newblock Model-based and model-free mechanisms of human motor learning.
\newblock In \emph{Progress in motor control}, pp.\  1--21. Springer, 2013.

\bibitem[Harmon et~al.(1995)Harmon, Baird~III, and Klopf]{harmon1995advantage}
Harmon, M.~E., Baird~III, L.~C., and Klopf, A.~H.
\newblock Advantage updating applied to a differential game.
\newblock In \emph{Advances in neural information processing systems}, pp.\
  353--360, 1995.

\bibitem[Kalashnikov et~al.(2018)Kalashnikov, Irpan, Pastor, Ibarz, Herzog,
  Jang, Quillen, Holly, Kalakrishnan, Vanhoucke, et~al.]{kalashnikov2018qt}
Kalashnikov, D., Irpan, A., Pastor, P., Ibarz, J., Herzog, A., Jang, E.,
  Quillen, D., Holly, E., Kalakrishnan, M., Vanhoucke, V., et~al.
\newblock Qt-opt: Scalable deep reinforcement learning for vision-based robotic
  manipulation.
\newblock \emph{arXiv preprint arXiv:1806.10293}, 2018.

\bibitem[Kappen(2005)]{kappen2005path}
Kappen, H.~J.
\newblock Path integrals and symmetry breaking for optimal control theory.
\newblock \emph{Journal of statistical mechanics: theory and experiment},
  2005\penalty0 (11):\penalty0 P11011, 2005.

\bibitem[Lillicrap et~al.(2015)Lillicrap, Hunt, Pritzel, Heess, Erez, Tassa,
  Silver, and Wierstra]{lillicrap2015continuous}
Lillicrap, T.~P., Hunt, J.~J., Pritzel, A., Heess, N., Erez, T., Tassa, Y.,
  Silver, D., and Wierstra, D.
\newblock Continuous control with deep reinforcement learning.
\newblock \emph{arXiv preprint arXiv:1509.02971}, 2015.

\bibitem[Mnih et~al.(2015)Mnih, Kavukcuoglu, Silver, Rusu, Veness, Bellemare,
  Graves, Riedmiller, Fidjeland, Ostrovski, et~al.]{mnih2015human}
Mnih, V., Kavukcuoglu, K., Silver, D., Rusu, A.~A., Veness, J., Bellemare,
  M.~G., Graves, A., Riedmiller, M., Fidjeland, A.~K., Ostrovski, G., et~al.
\newblock Human-level control through deep reinforcement learning.
\newblock \emph{Nature}, 518\penalty0 (7540):\penalty0 529, 2015.

\bibitem[Murphy(2012)]{Murphy2012}
Murphy, K.~P.
\newblock \emph{Machine Learning: A Probabilistic Perspective}.
\newblock The MIT Press, 2012.
\newblock ISBN 0262018020, 9780262018029.

\bibitem[Pan et~al.(2015)Pan, Theodorou, and Kontitsis]{pan2015sample}
Pan, Y., Theodorou, E., and Kontitsis, M.
\newblock Sample efficient path integral control under uncertainty.
\newblock In \emph{Advances in Neural Information Processing Systems}, pp.\
  2314--2322, 2015.

\bibitem[Precup et~al.(1998)Precup, Sutton, and Singh]{precup1998theoretical}
Precup, D., Sutton, R.~S., and Singh, S.
\newblock Theoretical results on reinforcement learning with temporally
  abstract options.
\newblock In \emph{European conference on machine learning}, pp.\  382--393.
  Springer, 1998.

\bibitem[Saxe et~al.(2017)Saxe, Earle, and Rosman]{saxe2017hierarchy}
Saxe, A.~M., Earle, A.~C., and Rosman, B.
\newblock Hierarchy through composition with multitask {LMDP}s.
\newblock In Precup, D. and Teh, Y.~W. (eds.), \emph{Proceedings of the 34th
  International Conference on Machine Learning}, volume~70 of \emph{Proceedings
  of Machine Learning Research}, pp.\  3017--3026, International Convention
  Centre, Sydney, Australia, 06--11 Aug 2017. PMLR.
\newblock URL \url{http://proceedings.mlr.press/v70/saxe17a.html}.

\bibitem[Schaul et~al.(2015)Schaul, Horgan, Gregor, and
  Silver]{schaul2015universal}
Schaul, T., Horgan, D., Gregor, K., and Silver, D.
\newblock Universal value function approximators.
\newblock In \emph{International Conference on Machine Learning}, pp.\
  1312--1320, 2015.

\bibitem[Schrempf et~al.(2005)Schrempf, Feiermann, and
  Hanebeck]{schrempf2005optimal}
Schrempf, O.~C., Feiermann, O., and Hanebeck, U.~D.
\newblock Optimal mixture approximation of the product of mixtures.
\newblock In \emph{International Conference on Information Fusion}, volume~1,
  pp.\  8--pp. IEEE, 2005.

\bibitem[Silver et~al.(2016)Silver, Huang, Maddison, Guez, Sifre, Van
  Den~Driessche, Schrittwieser, Antonoglou, Panneershelvam, Lanctot,
  et~al.]{silver2016mastering}
Silver, D., Huang, A., Maddison, C.~J., Guez, A., Sifre, L., Van Den~Driessche,
  G., Schrittwieser, J., Antonoglou, I., Panneershelvam, V., Lanctot, M.,
  et~al.
\newblock Mastering the game of go with deep neural networks and tree search.
\newblock \emph{Nature}, 529\penalty0 (7587):\penalty0 484, 2016.

\bibitem[Tassa et~al.(2018)Tassa, Doron, Muldal, Erez, Li, Casas, Budden,
  Abdolmaleki, Merel, Lefrancq, et~al.]{tassa2018deepmind}
Tassa, Y., Doron, Y., Muldal, A., Erez, T., Li, Y., Casas, D. d.~L., Budden,
  D., Abdolmaleki, A., Merel, J., Lefrancq, A., et~al.
\newblock Deepmind control suite.
\newblock \emph{arXiv preprint arXiv:1801.00690}, 2018.

\bibitem[Todorov(2009)]{todorov2009compositionality}
Todorov, E.
\newblock Compositionality of optimal control laws.
\newblock In \emph{Advances in Neural Information Processing Systems}, pp.\
  1856--1864, 2009.

\bibitem[Todorov et~al.(2012)Todorov, Erez, and Tassa]{todorov2012mujoco}
Todorov, E., Erez, T., and Tassa, Y.
\newblock Mujoco: A physics engine for model-based control.
\newblock In \emph{Intelligent Robots and Systems (IROS)}, pp.\  5026--5033.
  IEEE, 2012.

\bibitem[van Niekerk et~al.(2018)van Niekerk, James, Earle, and
  Rosman]{van2018will}
van Niekerk, B., James, S., Earle, A., and Rosman, B.
\newblock Will it blend? composing value functions in reinforcement learning.
\newblock \emph{arXiv preprint arXiv:1807.04439}, 2018.

\bibitem[Wang et~al.(2015)Wang, Schaul, Hessel, Van~Hasselt, Lanctot, and
  De~Freitas]{wang2015dueling}
Wang, Z., Schaul, T., Hessel, M., Van~Hasselt, H., Lanctot, M., and De~Freitas,
  N.
\newblock Dueling network architectures for deep reinforcement learning.
\newblock \emph{arXiv preprint arXiv:1511.06581}, 2015.

\bibitem[Ziebart et~al.(2008)Ziebart, Maas, Bagnell, and
  Dey]{ziebart2008maximum}
Ziebart, B.~D., Maas, A.~L., Bagnell, J.~A., and Dey, A.~K.
\newblock Maximum entropy inverse reinforcement learning.
\newblock In \emph{AAAI}, volume~8, pp.\  1433--1438. Chicago, IL, USA, 2008.

\end{thebibliography}
\bibliographystyle{icml2019}

\onecolumn
\appendix
\section{Proofs} \label{appendix:proofs}

\subsection{Max-Ent Generalized Policy Improvement}

\label{appendix:maxentgpi}
\maxentgpi*

For brevity we denote $Q^{\max} \definedas \max_i Q^i$. Define the soft Bellman operator associated with policy $\pi$ as
\begin{align*}
    \mathcal{T}^\pi Q(s, a)
     & \definedas r(s, a, s') + \gamma \E_{p(s'|s, a)}\left[ \alpha H[\pi(\cdot|s')] + \E_{a' \sim \pi(\cdot|s')}\left[ Q(s', a') \right] \right].
\end{align*}
\citet{haarnoja2018soft} have pointed out that the soft Bellman operator $\mathcal{T}^\pi$ corresponds to a conventional, ``hard'', Bellman operator defined over the same MDP but with reward $r_{\pi}(s,a,s') = r(s, a, s') + \gamma \alpha \E_{p(s'|s, a)}\left[ H[\pi(\cdot|s')]\right]$. Thus, as long as $r(s,a,s')$ and $H[\pi(\cdot|s')]$ are bounded, $\mathcal{T}^\pi$ is a contraction with $Q^\pi$ as its fixed point.
Applying $\mathcal{T}^\pi$ to $Q^{\max}(s, a)$ we have:
\begin{align*}
    \mathcal{T}^\pi Q^{\max}(s, a)
    \nonumber & = r(s, a, s') + \gamma \E_{s' \sim p(\cdot |s, a), a' \sim \pi(\cdot|s')}\left[ -\alpha \log \pi(a'|s') + Q^{\max}(s', a') \right]\\
    \nonumber & = r(s, a, s') + \gamma \E_{s' \sim p(\cdot |s, a), a' \sim \pi(\cdot|s')}\left[ -\alpha \log \dfrac{\exp (\alpha^{-1} Q^{\max}(s',a'))}{Z^\pi(s')} + Q^{\max}(s', a') \right] \\
    \nonumber & = r(s, a, s') + \gamma \E_{s' \sim p(\cdot |s, a)} \left[ \alpha \log Z^{\pi}(s') \right].
    \end{align*}
Similarly, if we apply $\mathcal{T}^{\pi_i}$, the soft Bellman operator induced by policy $\pi_i$, to $Q^{\max}(s, a)$, we obtain:
\begin{align*}
    \mathcal{T}^{\pi_i} Q^{\max}(s, a)
    \nonumber & = r(s, a, s') + \gamma \E_{s' \sim p(\cdot |s, a), a' \sim \pi_i(\cdot|s')}\left[ -\alpha \log \pi_i(a'|s') + Q^{\max}(s', a') \right].
\end{align*}

We now note that the Kullback-Leibler divergence between $\pi_i$ and $\pi$ can be written as
\begin{align*}
    \KL({\pi_i(\cdot|s)}\|{\pi(\cdot|s)})
    \nonumber & = \E_{a \sim \pi_i(\cdot|s)}\left[ \log \pi_i(a|s) - \log \pi(a|s) \right] \\
    \nonumber & = \E_{a \sim \pi_i(\cdot|s)}\left[ \log \pi_i(a|s) - \frac{1}{\alpha} Q^{\max}(s, a) + \log Z^\pi(s) \right].
\end{align*}
The quantity above, which is always nonnegative, will be useful in the subsequent derivations. Next we write
\begin{align}
\label{eq:pige}
\mathcal{T}^\pi Q^{\max}(s, a) - \mathcal{T}^{\pi_i} Q^{\max}(s, a)
\nonumber & = \gamma \E_{s' \sim p(\cdot |s, a)}\left[ \alpha \log Z^{\pi}(s') - \E_{a' \sim \pi_i(\cdot|s')}[-\alpha \log \pi_i(a'|s') + Q^{\max}(s', a')] \right]\\
\nonumber & = \gamma \E_{s' \sim p(\cdot |s, a)}\left[ \E_{a' \sim \pi_i(\cdot|s')}[\alpha \log Z^{\pi}(s') +\alpha \log \pi_i(a'|s') -Q^{\max}(s', a')] \right]\\
\nonumber & = \gamma \E_{s' \sim p(\cdot |s, a)}\left[  \alpha \KL({\pi_i(\cdot|s')}\|{\pi(\cdot|s')}) \right] \\
& \ge 0.
\end{align}

From~(\ref{eq:pige}) we have that
\begin{align*}
    \mathcal{T}^\pi Q^{\max}(s, a) \geq \mathcal{T}^{\pi_i} Q^{\max}(s, a) \geq \mathcal{T}^{\pi_i} Q^i(s, a) = Q^i(s, a) \; \text{ for all } i.
\end{align*}
Using the contraction and monotonicity of the soft Bellman operator $\mathcal{T}^\pi$ we have
\begin{align*}
    Q^\pi(s, a) &= \lim_{k \rightarrow \infty} (\mathcal{T}^\pi)^k Q^{\max}(s, a)
    \geq Q^i(s, a) \; \text{ for all } i.
\end{align*}
We have just showed~(\ref{eq:gpiQ}). In order to show~(\ref{eq:gpiV}), we note that
\begin{align}
\label{eq:vpilogz}
    V^\pi(s)
    \nonumber &\definedas  \alpha H[\pi(\cdot|s)] + \E_{a\sim \pi}\left[ Q^\pi(s, a) \right] \\
\nonumber & \geq  \alpha H[\pi(\cdot|s)] + \E_{a\sim \pi}\left[ Q^{\max}(s, a) \right]
    \\
    &= \alpha \log Z^{\pi}(s).
\end{align}
Similarly, we have, for all $i$,
\begin{align}
\label{eq:vilogz}
V^i(s)
\nonumber & = \E_{a \sim \pi_i(\cdot|s)} \left[Q^i(s,a) - \alpha \log \pi_i(a|s) \right] \\
\nonumber & \le \E_{a \sim \pi_i(\cdot|s)} \left[Q^{\max}(s,a) - \alpha \log \pi_i(a|s) \right] \\
\nonumber & = \alpha \log Z^\pi(s) -\alpha \KL({\pi_i(\cdot|s)}\|{\pi(\cdot|s)}) \\
& \le \alpha \log Z^\pi(s).
\end{align}
The bound~(\ref{eq:gpiV}) follows from~(\ref{eq:vpilogz}) and~(\ref{eq:vilogz}).

\subsection{DC Proof} \label{appendix:ltd_proof}

\ltdtheorem*

We follow a similar approach to \cite{haarnoja2018composable} but without making approximations and generalizing to all convex combinations.

First note that since $\pi_i$ and $\pi_j$ are optimal then $\pi_i(a|s) = \exp({\frac{1}{\alpha}(Q^i(s,a) - V^i(s))})$.

For brevity we use $s$ and $s'$ notation rather than writing the time index.

Define
\begin{align}
    Q^{(0)}_b(s, a) &\definedas b Q^i(s, a) + (1 - b)Q^j(s, a)
    \\
    C^{(0)}(s, a) &\definedas 0
\end{align}

and consider soft Q-iteration on $r_b$ starting from $Q^{(0)}_b$. We prove, inductively, that at each iteration $Q^{(k+1)}_b = b Q^i(s, a) + (1 - b)Q^j(s, a) - C^{(k+1)}(s, a)$.

This is true by definition for $k = 0$.
\begin{align}
    Q_b^{(k+1)}(s, a) &= r_b(s, a) + \gamma \alpha \E_{p(s'|s, a)}\left[ \log \int_\mathcal{A} \exp \frac{1}{\alpha}Q_b^{(k)}(s', a')  da' \right]
    \\
    &= r_b(s, a) + \\ \notag  &\gamma \alpha \E_{p(s'|s, a)}\left[ \log \int_\mathcal{A} \exp(\frac{1}{\alpha}(b Q^i(s', a') + (1 - b) Q^j(s', a') -  C^{(k)}(s', a')) )da' \right]
    \\
    &= r_b(s, a) + \\ \notag
    &\E_{p(s'|s, a)} \left[ b V^i(s') + (1 - b) V^j(s') + \alpha \log \int_{\mathcal{A}} \exp( b \log \pi_i(a'|s') + (1 - b) \log \pi_j(a'|s') - \frac{1}{\alpha}C^{(k)}(s', a')) da' \right]
    \\
    &= b Q^i(s, a) + (1 - b) Q^j(s, a) + \\ \notag & \alpha \gamma \E_{p(s'|s, a)} \left[\log \int_{\mathcal{A}} \exp( b \log \pi_1(a'|s') + (1 - b) \log \pi_2(a'|s') - \frac{1}{\alpha}C^{(k)}(s', a')) da' \right]
    \\
    &= b Q^i(s, a) + (1 - b) Q^i(s, a) - C^{(k+1)}_b(s, a).
\end{align}

Since soft Q-iteration converges to the $\alpha$ max-ent optimal soft $Q$ then at the limit $k \rightarrow \inf$ theorem \ref{theorem:ltd}  holds.

One can get an intuition for $C^\infty_b(s,a)$ by noting that \begin{align}
    C^{(1)}_b(s, a) = \gamma\alpha \E_{p(s'|s, a)}\left[ (1 - b) \renyi{b}{\pi_1(\cdot|s)}{\pi_2(\cdot|s)}\right]
\end{align}
where $\renyiop_b$ is the R\'enyi divergence of order $b$. $C^\infty_b(s, a)$ can be seen as the discount sum of divergences, weighted by the unnormalized product distribution $\pi_1(a|s)^b \pi_2(a|s)^{1 - b}$.






\subsection{$N$ policies}

It is possible to extend Theorem 3.2 to the case with $N$ policies in a straightforward way.

\begin{restatable}[Multi-policy DC Optimality]{theorem}{ltdtheoremn} \label{theorem:ltdn}
    Let $\pi_1, \pi_2, ..., \pi_N$ be $\alpha$ max-ent optimal policies for tasks with rewards $r_1, r_2, ..., r_N$ with max-ent action-value functions $Q^1, Q^2, ..., Q^N$.

    Define $C^\infty_{\vect{w}}(s_t, a_t)$ as the fixed point of
    \begin{align*} \textstyle
        C^{(k+1)}_{\vect{w}}(s_t, a_t) = -\alpha \gamma \E_{p(s_{t+1}|s_t, a_t)} \left[ \log \int_\mathcal{A} \left(\prod_{i=1}^N \pi_i(a_{t+1}|s_{t+1})^{w_i } \right) \exp(-\frac{1}{\alpha} C^{(k)}_{\vect{w}}(s_{t+1}, a_{t+1})) da_{t+1} \right]
    \end{align*}
    Given the conditions for Soft Q convergence,
the max-ent optimal $Q^*_{\vect{w}}(s, a)$ for the convex combination of rewards $r_\vect{w} = \sum_{i=1}^N r_i w_i$ is
    \begin{align*} \label{eq:cn} \textstyle
        Q^*_{\vect{w}}(s, a) = \sum_{i=1}^N w_i Q^i(s, a) - C^\infty_{\vect{w}}(s, a)
        \\
    \forall s \in \mathcal{S}, a \in \mathcal{A}, \vect{w} \in \{ \vect{w} | \sum_{i=1}^N w_i = 1 \:\:\: \mathrm{and} \:\:\: w_i  \geq 0 \}
\end{align*}
\end{restatable}

Note that $w_i$ refers to component $i$ of the vector $\vect{w}_i$.

The proof is very similar to the two reward case above.

Define
\begin{align}
    Q_{\vect{w}}^{(0)} \definedas \sum_{i=1}^N w_i Q^i(s, a)
    \\
    C_{\vect{w}}^{(0)} \definedas 0
\end{align}

and again consider soft Q-iteration on $r_{\vect{w}}$. We prove by induction that at each iteration
\begin{align}
    Q^{(k+1)}_{\vect{w}}(s, a) = \sum_{i=1}^N w_i Q^i(s, a) - C_{\vect{w}}^{(k+1)}(s, a)
\end{align}
Again, this is true by definition for $k=0$. Now we consider a step of Soft Q iteration

\begin{align}
    Q_{\vect{w}}^{(k+1)} &= r_{\vect{w}}(s, a) + \gamma\alpha \E_{p(s'|s, a)}\left[ \log \int_{\mathcal{A}} \exp \frac{1}{\alpha} Q^{(k)}_{\vect{w}}(s', a') da' \right]
    \\
    &=  r_{\vect{w}}(s, a) + \gamma\alpha \E_{p(s'|s, a)}\left[ \log \int_{\mathcal{A}} \exp \frac{1}{\alpha} \left( \sum_{i=1}^N w_i Q^i(s', a')  - C_{\vect{w}}^{(k)}(s, a)\right) da' \right]
    \\
    &= r_{\vect{w}}(s, a) + \gamma \E_{p(s'|s, a)}\left[ \sum_{i=1}^N w_i V^i(s') + \alpha \log \int_{\mathcal{A}} \exp \left( \sum_{i=1}^N w_i \log \pi_i(a'|s') - \frac{1}{\alpha}C^{(k)}_{\vect{w}}(s', a') \right) da' \right]
    \\
    &= \sum_{i=1}^N w_i Q^i(s, a) + \alpha \gamma \E_{p(s'|s,a)}\left[ \log \int_{\mathcal{A}} \exp(\sum_{i=1}^N w_i \log \pi_i(a'|s') - \frac{1}{\alpha} C_{\vect{w}}^{(k)}(s', a')) da' \right]
    \\
    &= \sum_{i=1}^N w_i Q^i(s, a) - C_{\vect{w}}^{(k+1)}(s, a)
\end{align}

Since soft Q-iteration converges to the $\alpha$ max-ent optimal soft $Q$ then $Q^*_{\vect{w}}(s, a) = \sum_{i=1}^N w_i Q^i(s, a) - C^{(k+1)}_{\vect{w}}(s, a)$ for all $s \in \mathcal{S}, \: a \in \mathcal{A}$.

Note that, in practice, estimating $C^{\infty}_{\vect{w}}$ may be more challenging for larger $N$. For compositions of many policies, GPI may be more practical.

\section{Theoretical properties of the composition methods}

\begin{table}[H]
    \centering
    \begin{tabular}{p{2cm}|p{2.5cm}|p{2.5cm}|p{2cm}|p{2.8cm}}
        \textbf{Method} & \textbf{Optimal} & \textbf{Bounded loss} & \textbf{Requires $\sfr$} & \textbf{Requires $f(s, a | b)$}
        \\        \hline
         CO&
         & & &  \\ \hline
         CondQ & {\hfil \checkmark} & {\hfil na} & {\hfil \checkmark} & {\hfil \checkmark}  \\ \hline
         GPI & & {\hfil \checkmark} & {\hfil \checkmark} \\
        \hline
        DC & \hfil \checkmark & {\hfil na} &  & {\hfil \checkmark}\\
    \end{tabular}
    \caption{Theoretical properties of different approaches to max-ent transfer. The methods compared are: CO, CondQ, max-ent GPI (over a fixed, finite set of policies), and DC. The columns indicate whether the transfer policy is optimal, the regret of the transfer policy is bounded, whether rewards for all tasks $\sfr$ need to be observed simultaneously during training and whether the method requires learning a function conditional on the transfer task $b$, $f(s, a|b)$.
    DC is the only method that both recovers (in principle) the optimal policy and does not require observing $\phi$ during training.}
    \label{tab:comparison}
\end{table}

\section{Algorithm details}
\label{appendix:algo_details}

\subsection{Transfer algorithm}
\begin{algorithm}[h]
    \begin{algorithmic}
        \State Load trained parameters $\theta_Q$, $\theta_q$, $\theta_\sfQ$, $\theta_C$, $\theta_{Q_b}$.
        \State Accept transfer task parameter $b$, transfer $method \in$ {CO, GPI, DC, CondQ}.
        \While{testing}
            \State Importance sample transfer policy $\pi_b(a|s) \propto \exp \frac{1}{\alpha} Q^{method}(s, a)$ with mixture proposal $p_{b}(a|s)_{\theta_q}$.
        \EndWhile
    \end{algorithmic}
    \caption{AISBP transfer algorithm \label{algo:transfer}}
\end{algorithm}

\subsection{All losses and estimators}

We use neural networks to parametrize all quantities. For each policy we learn an action-value $Q_{\theta_Q}(s, a)$, value $V_{\theta_V}(s)$ and proposal distribution $q_{\theta_q}(a|s)$. We use target networks for the proposal distribution $q_{\theta_q'}(a|s)$ and value $V_{\theta_V'}(s)$.

Here we enumerate all of the losses and their estimators. We use temporal difference (TD(0)) learning for all the RL losses, so all losses are valid off-policy. We use a replay buffer $R$ and learn by sampling minibatches of SARS tuples of size $B$, we index over the batch dimension with $l$ and use $s'_l$ to denote the state following $s_l$, so the tuple consists of $(s_l, a_l, r_l, s_l')$. For importance sampled estimators we sample $N$ actions for each state $s_l$ and use $a_{lk}$ to denote sample $k$ for state $l$.

We learn a set of $n$  policies, one for each task in $\mathcal{T}$ indexed by $i$. However, we write the losses for a single policy and drop $i$ for notational simplicity.

\subsubsection{Proposal loss}
The proposal loss minimizes the KL divergence between the Boltzmann distribution $\pi(a|s) \propto \exp(\frac{1}{2}Q(s, a))$ and the proposal distribution.

\begin{align}
    \mathcal{L}(\theta_q) &= \E_{R}\left[ \E_{a \sim \pi(\cdot|s)}[\log \pi(a|s_t) - \log q_{\theta_q}(a|s_t)] \right]
\end{align}

As described in the text, this loss is estimated using importance sampling with a mixture distribution $p(a|s)$ containing equally weighted components consisting of the target proposal distribution $q_{\theta_q'}(a|s)$ for all policies and the uniform distribution.

\begin{align} \label{eq:mixture}
    p(a|s) = \frac{1}{n+1}\left(\frac{1}{V^\mathcal{A}} + \sum_{i=1}^n q^i_{\theta_q'}(a|s)\right)
\end{align}
where $V^{\mathcal{A}}$ is the volume of the action space (which is always bounded in our case).

The proposal loss is estimated using self-normalized importance sampling
\begin{align}
    \mathcal{L}(\theta_q) & \approx -\frac{1}{B }\sum_{k=1}^{B} \sum_{l=1}^N w_{kl} \log q_{\theta_q}(a|s_t),
    \\
    w'_{kl} &= \frac{\frac{1}{\alpha}(Q_{\theta_Q}(s_k, a_{kl})) }{p(a_{kl}| s_k)} ; \: \: w_{kl} = \frac{w_{kl'}}{\sum_{m=1}^N w'_{km}}. \label{eq:proposal_update}
\end{align}

\subsubsection{Value loss}

The soft value loss is
\begin{align} \textstyle
    \mathcal{L}(\theta_V) = & \E_{R}\left[ \frac{1}{2}(V_{\theta_V}(s_t) - \alpha \log \int_{\mathcal{A}} \exp(\frac{1}{\alpha}Q_{\theta_Q}(s_t, a)) da )^2 \right] \label{eq:valueloss2}
\end{align}

We estimate this using importance sampling with the proposal distribution $q_{\theta_q}(a|s)$ which is trying to fit the policy $\pi$.

\begin{align}
    \mathcal{L}(\theta_V) &\approx \frac{1}{2B}\sum_{l=1}^B \left(V_{\theta_V}(s_l) - \alpha \log Z \right)^2
    \\
    Z &= \left[\frac{1}{N} \sum_{k=1}^N \frac{\exp(\frac{1}{\alpha} Q_{\theta_Q}(s_{l}, a_{lk}))}{q_{\theta_q}(a_{lk}|s_l)} \right]  \label{eq:alpha_logz_est_appendix}
\end{align}

\subsubsection{Action-value loss}

The TD(0) loss for $Q_{\theta_Q}$ is

\begin{align}
    \mathcal{L}(\theta_Q) =& \E_{R}\left[\frac{1}{2}(Q_{\theta_Q}(s_t, a_t) - (r(s_t, a_t, s_{t+1}) + \gamma V_{\theta_V'}(s_{t+1})))^2 \right] \label{eq:actionvalueloss}
\end{align}

This does not require importance sampling to estimate and can be straightforwardly estimated as
\begin{align}
    \mathcal{L}(\theta_Q) \approx \frac{1}{2B}\sum_{l =1}^B (Q_{\theta_Q}(s_l, a_l) - (r_l + \gamma V_{\theta_V'}(s')))^2
\end{align}

The action-value is parametrized as an advantage function $Q_{\theta_Q}(s, a) = V_{\theta_V'}(s) + A_{\theta_A}(s, a)$.

\subsubsection{State Dependent Successor Features loss}

To facilitate max-ent GPI we learn successor features for each policy, both state-action dependent features $\sfQ_{\theta_\sfQ}(s, a)$ and state-dependent $\sfV_{\theta_\sfV}(s)$. As with value, we use a target network for the state-dependent features $\sfV_{\theta_{\sfV}'}(s)$

\begin{align*} \textstyle
    \mathcal{L}(\theta_\sfV) = & \E_{R}\left[ \frac{1}{2}(\sfV_{\theta_\sfV}(s_t) - \E_{a_t \sim \pi(a_t|s_t)}[\sfQ_{\theta_\sfQ}(s_t, a_t) + \alpha {\bm{1}}(-Q_{\theta_Q}(s_t, a_t) + \alpha \log Z(s_t))])^2 \right] \label{eq:sfvalueloss}
\end{align*}

This loss is estimated using self-normalized importance sampling with proposal $q_{\theta_q}$
\begin{align}
    \mathcal{L}(\theta_{\sfV}) &\approx \frac{1}{2B}\sum_{l=1}^B \sum_{k=1}^N w_{lk} \left[(\sfQ_{\theta_{\sfQ}}^i(s_l, a_{lk}) - Q^i_{\theta_{Q}}(s_l, a_{lk}) + \alpha \log Z(s_l))^2\right],
    \\
    w_{lk} & \propto \frac{\exp(\frac{1}{\alpha} Q^i(s_l, a_lk)) }{q^i_{\theta_q}(a_{lk}|s_l)}.
\end{align}
We use the importance sampled estimate of $Z$ from eq \ref{eq:alpha_logz_est_appendix}, rather than the value network which may be lagging the true partition function. We use self-normalized importance sampling to avoid the importance weights depending on $\alpha \log Z(s_l)$ (this introduces a bias, but in practise appears to work well).

\subsubsection{State-Action Dependent Successor Features loss}

The state-action dependent successor feature loss is
\begin{align}
        \mathcal{L}(\theta_\sfQ) =& \E_{R}\left[\frac{1}{2}(\sfQ_{\theta_Q}(s_t, a_t) - (\sfr(s_t, a_t, s_{t+1}) + \gamma \sfV_{\theta_\sfV'}(s_{t+1})))^2 \right]. \label{eq:sfloss}
\end{align}

for which we use the following estimator
\begin{align}
    \mathcal{L}(\theta_{\sfQ^i}) \approx \frac{1}{2B}\sum_{l =1}^B (\sfQ^i_{\theta_\sfQ}(s_l, a_l) - (\sfr_l + \gamma \sfV_{\theta_\sfV'}(s_l')))^2.
\end{align}

$\sfQ_{\theta_{\sfQ}}$ is parametrized as a ``psi-vantage'' $\sfQ_{\theta_{\sfQ}}(s, a) = \sfV_{\theta_\sfV'}(s) + \sfQ_{\theta_A}^A(s, a)$.

\subsubsection{DC correction}

We learn the divergence correction for each pair of policies $\pi_i(a|s)$, $\pi_j(a|s)$. As described in the text, in order to learn $C_{\theta_C}(s, a, b)$ for all $b \in [0, 1]$, we sample $b$. We also use a target network $C_{\theta_C'}(s, a, b)$. The loss is then
\begin{align} \textstyle
    \mathcal{L}(\theta_C) = \E_{s \sim R, b \sim U(0, 1)} [ \frac{1}{2} (
    C_{\theta_C}(s, a, b) + \alpha \gamma \E_{p(s'|s, a)}[ \log \int_{\mathcal{A}} \exp(b \log \pi_i(a'|s') + \\ \notag (1 - b) \pi_j(a'|s') - \frac{1}{\alpha}C_{\theta_{C'}}(s', a', b)) da' ]
    )^2 ].
\end{align}

This loss is challenging to estimate, due to the dependence on two policies. We importance sample using a mixture of all proposal distributions uniform $p(a|s)$ (equation \ref{eq:mixture}). We denote the samples of $b \sim \mathcal{U}(0, 1)$ for each batch entry $b_l$. Note the choice of uniform distribution for $b$ is not required, other distributions that ensure the estimator works well for $b \in [0, 1]$ would also work. The importance sampled estimator is then

\begin{align}
    \mathcal{L}(\theta_C) &\approx \frac{1}{N} \sum_{l=1}^B\left(C_{\theta_C}(s_l, a_{l}, b_l) - \alpha \gamma \log \left[ \frac{1}{N} \sum_{k=1}^N \frac{ C^{target}_{\theta_C'}(s_l', a_{lk}', b_l)}{p(a_{lk'}|s_l)} \right]  \right)^2,
    \\
    C^{target}_{\theta_C'}(s_l', a_{lk}', b_l) &\definedas \exp(\frac{1}{\alpha}(b_l Q^i_{\theta_Q}(s_l', a_{lk}') + (1 - b_l) Q^j_{\theta_Q}(s_l', a_{lk}') - C_{\theta_C'}(s_l', a_{lk}', b_l)).
\end{align}

We parametrized $C_{\theta_C}$ as an advantage function $C_{\theta_C}(s, a, b) = C_{\theta_{C^A}}^A(s, a, b) + C_{\theta_{C^B}}^B(s, b)$ with an additional loss to constrain this parametrization
\begin{align}
    \mathcal{L}(\theta_B) = \E_{a \sim q(\cdot|s), s \sim R}\left[ \frac{1}{2}(C_{\theta_{C^A}}^A(s, a, b))^2 \right]
\end{align}
which can be straightforwardly estimated by sampling from $q$
\begin{align}
    \mathcal{L}(\theta_B) \approx \frac{1}{2NB} \sum_{l=1}^B \sum_{k=1}^N (C_{\theta_{C^A}}^A(s_l, a_{lk}, b_l))^2
\end{align}

\subsubsection{CondQ}

We also consider, as a control, learning the action-value function conditional on $b$ directly \citep{schaul2015universal}, in a similar way to the DC correction.
We learn both a conditional value $V_{\theta_{V_b}}(s, b)$ and $Q_{\theta_{Q_b}}(s, a, b)$, again by sampling $b$ uniformly each update.
\begin{align} \textstyle
    \mathcal{L}(\theta_{V_b}) &= \E_{R, b \sim U(0, 1)}\left[ \frac{1}{2}(V_{\theta_{V_b}}(s, b) - \alpha \log \int \exp(\frac{1}{\alpha} Q_{\theta_{Q_b}}(s, a, b)))^2 \right],
    \\
    \mathcal{L}{\theta_Q} &= \E_{R, b \sim U(0, 1)}\left[ \frac{1}{2}(Q_{\theta_{Q_b}}(s, a, b) - (r_b + \gamma V_{\theta_{V_b}}(s', b)))^2\right],
\end{align}
where computing $r_b$ for arbitrary $b$ requires $\phi$ to have been observed.

We estimate Cond-Q with the same importance samples as $C$ from $p(a|s)$ and again sample $b \sim \mathcal{U}(0, 1)$ for each entry in the batch.
We use target networks for $V_{\theta_V'}(s, b)$ and parametrize $Q_{\theta_Q}(s, a, b) = V_{\theta_V'}(s, b) + A_{\theta_A}(s, a, b)$. 

The conditional value estimator is
\begin{align}
    \mathcal{L}(\theta_V) &\approx \frac{1}{2B} \sum_{l=1}^B \left(V_{\theta_{V_b}}(s_l, b_l) - \alpha \log \frac{1}{N}\sum_{k=1}^N \frac{\exp(\frac{1}{\alpha}Q_{\theta_{Q_b}}(s_l, a_{lk}, b_l)}{p(a_{lk}|s_l)}\right)^2
\end{align}
and action-value estimator is
\begin{align}
    \mathcal{L}(\theta_Q) & \approx \frac{1}{2B} \sum_{l=1}^B \left(Q_{\theta_{Q_b}}(s_l, a_l, b_l) - (r_b + \gamma V_{\theta_{V_b}'}(s_l', b_l)) \right)^2
\end{align}

\subsection{Sampling the product of proposals}
\label{subsec:transfer_sampling}

The proposal distributions $q^i(a|s)$ are mixtures of $M$ (truncated) normals (equation \ref{eq:mixture_of_normal}). We ignore the truncation when computing the product of proposals $q^{ij}(a|s)$.

The product of two $M$ component mixtures of normals results in another mixture of normals with $M^2$ components \citep[e.g.][]{schrempf2005optimal}. Since for all experiments $M$ is a relatively small integer (maximum is 16) we sample from the product of proposals in a naive way.

\section{Justification for the DC-Cheap heuristic} \label{appendix:ltd_heuristic}

We wish to estimate $C_b^\infty(s, a)$ (defined in Theorem \ref{theorem:ltd}) while avoiding learning a conditional function of $b$. We make two (substantial) assumptions to arrive at this approximation.

Firstly, we assume policies $\pi_i(a|s), \pi_j(a|s)$ are Gaussian
\begin{align} \label{eq:renyigaussian}
    \pi_i(a|s) = \exp\left(-\frac{(a - \mu_i(s))^2}{2 \sigma(s)^2}\right)
\end{align}
and the variance $\sigma(s)$ is the same for both policies given a state (it may vary across states).

Secondly, we assume $C_b^{(k)}(s, a) = C_b^{(k)}(s)$ is independent of action. This is approximately correct when nearby states have similar R\'enyi divergences between policies.

We make use of a result by \citet{gil2013renyi} that states that the R\'enyi divergence of order $b$ for two Gaussians of the same variance is
\begin{align}
   \renyi{b}{\mathcal{N}(\mu_1, \sigma)}{\mathcal{N}(\mu_2, \sigma)} = \frac{1}{2} \frac{b(\mu_1 - \mu_2)^2}{\sigma^2}.
\end{align}

We first define
\begin{align}
 G_b(s) \definedas (1 - b) \renyi{b}{\pi_i(\cdot|s)}{\pi_j(\cdot|s)} = -\log \int \pi_i(a|s)^b \pi_j(a|s)^{(1 - b)} da.
\end{align}
From \eqref{eq:renyigaussian}
\begin{align}
    G_b(s) = 4 b(1-b) G_{\frac{1}{2}}(s).
\end{align}

Given these assumptions we show inductively that $C^{(k)}_b(s, a) = 4 b (1 - b) C^{(k)}_{1/2}(s, a) \: \forall k, b \in [0, 1]$.

Since $C^{(0)}_b(s, a) = 0 \: \forall b \in [0, 1], a \in \mathcal{A}, s \in \mathcal{S}$ this is true for $k=0$. We show it holds inductively
\begin{align}
    C^{(k+1)}_b(s, a) &= -\alpha \gamma \E_{p(s'|s, a)}\left[ \log \int_{\mathcal{A}} \pi_i(a'|s')^b \pi_j(a'|s')^{(1-b)} \exp(-\frac{1}{\alpha} C_b^{(k)}(s', a')) da' \right]
    \\
    &=  \gamma \E_{p(s'|s, a)}\left[ \alpha G_b(s') + C_b^{(k)}(s') \right]
    \\
    &= 4 b (1 - b) C_{\frac{1}{2}}^{(k+1)}(s, a).
\end{align}


Obviously these assumptions are not justified. However, note that we estimate the true divergence for $C^{\infty}_{1/2}$, i.e. without any assumptions of Gaussian policies and this heuristic is used to estimate $C^{\infty}_b$ from $C^{\infty}_{1/2}$. In practise, we find this heuristic works in many situations where the policies have similar variance, particulary when bounded by GPI.

\section{Additional Figures}

\begin{figure}[H]
    \centering
    \includegraphics[width=\textwidth]{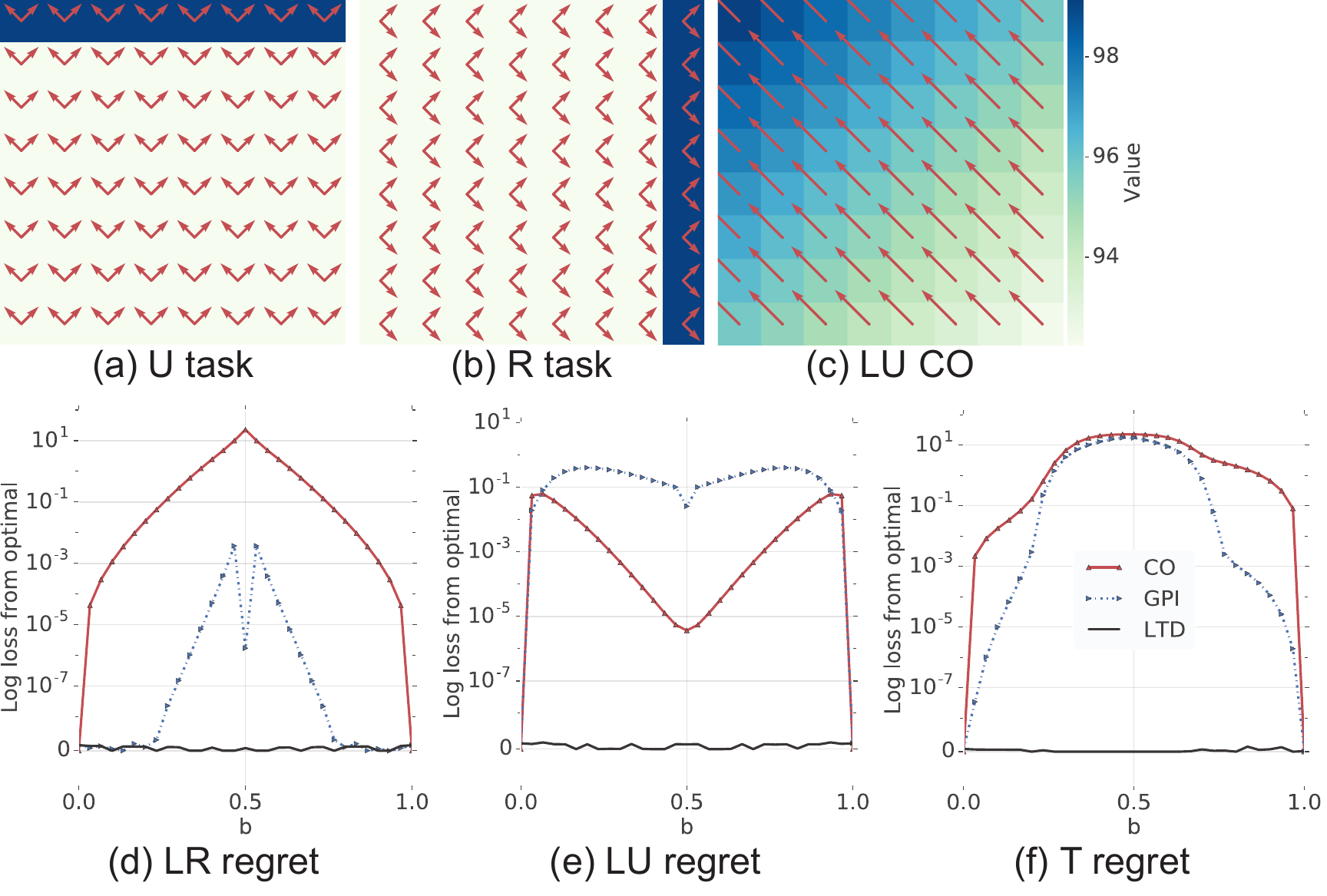}
    \caption{\textbf{Additional results for figure \ref{fig:tabular} (tabular)}
    \\
    (\textbf{a}) The U(p) and (\textbf{b}) R(ight) tasks.
    \newline
    (\textbf{c}) The CO policy for the LU task. Note how even far from the reward (e.g.\ bottom right corner) the CO policy is near optimal, contrast with the GPI policy for this task (figure 1f).
    \newline
    The log regret (smaller is better) as function of $b$ ($r_b = b r_1 + (1-b)r_2$) for the transfer task for the (\textbf{d}) incompatible (Left-Right) task, (\textbf{e}) compatible (Left Up) task and (\textbf{f}) T(ricky) task.
    \label{fig:tabular_supp}}
\end{figure}

\begin{figure}[H]
    \centering
    \includegraphics[width=\textwidth]{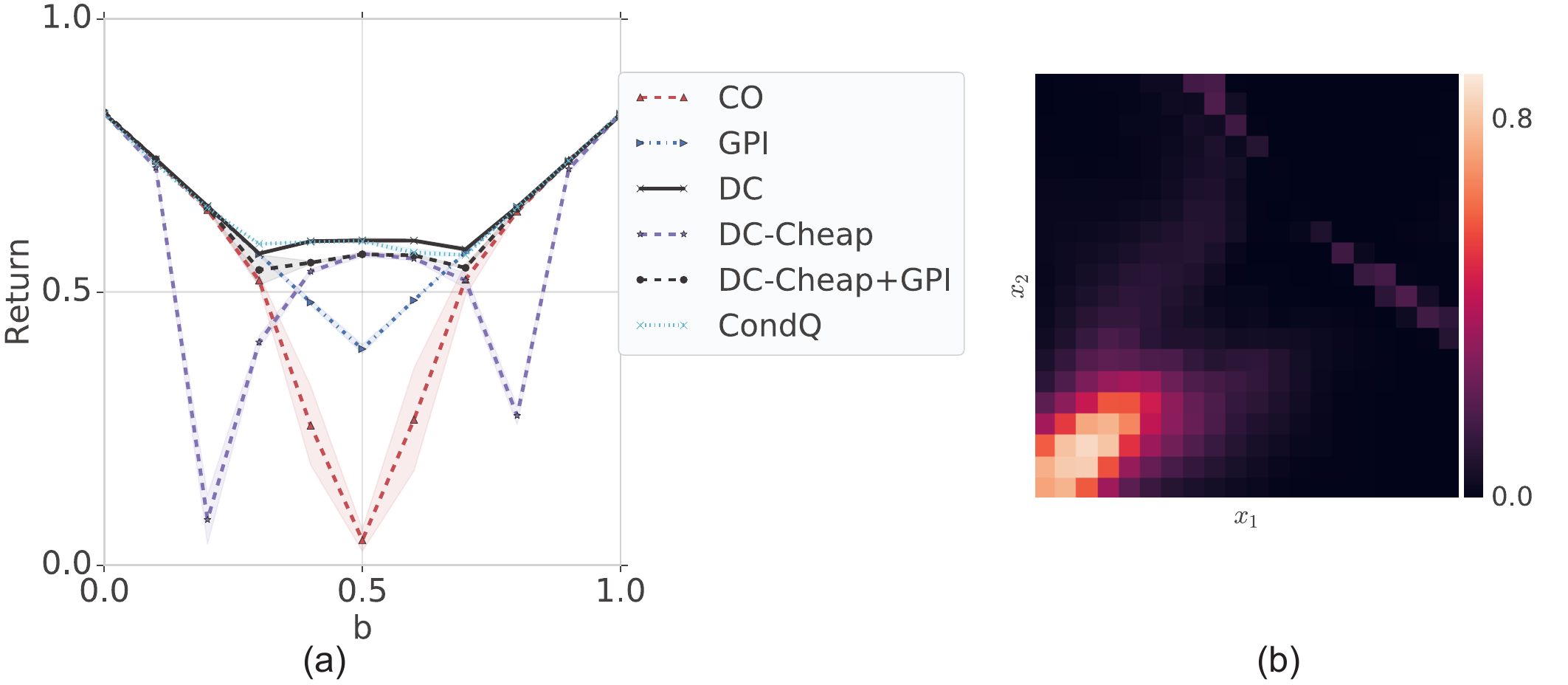}
    \caption{\textbf{Additional results for figure \ref{fig:point_mass} (point mass tricky)}
    \\
    (\textbf{a}) The returns (larger is better) for the transfer task as a function of $b$ ($r_b = b r_1 + (1-b)r_2$) including the DC heuristics. DC-Cheap+GPI performs almost as well as DC.
    \\
    (\textbf{b}) The R\'enyi divergence of the two base policies as a function of position: the two policies are compatible except near the bottom left corner where the rewards are non-overlapping.
    \label{fig:point_mass_sup}
    }
\end{figure}

\begin{figure}[h!]
    \centering
    \includegraphics[width=0.6\textwidth]{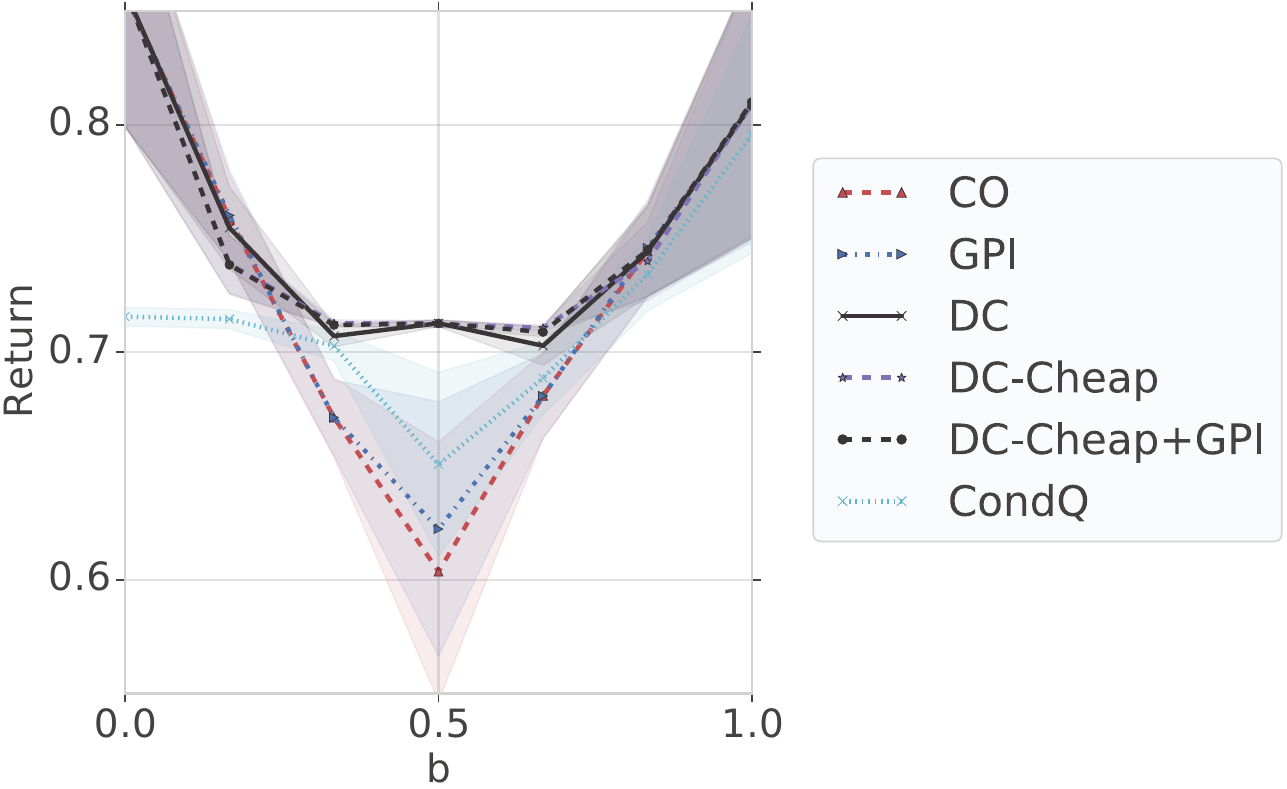}
    \caption{\textbf{Returns for figure \ref{fig:reacher} (planar manipulator)}
    \\
    The returns for the transfer task as a function of $b$ ($r_b = b r_1 + (1-b)r_2$) including the DC heuristics. DC-Cheap+GPI performs almost as well as DC. Shaded bars show SEM (5 seeds).}
    \label{fig:planar_manipulator_supp}
\end{figure}

\begin{figure}[H]
    \centering
    \includegraphics[width=0.7\textwidth]{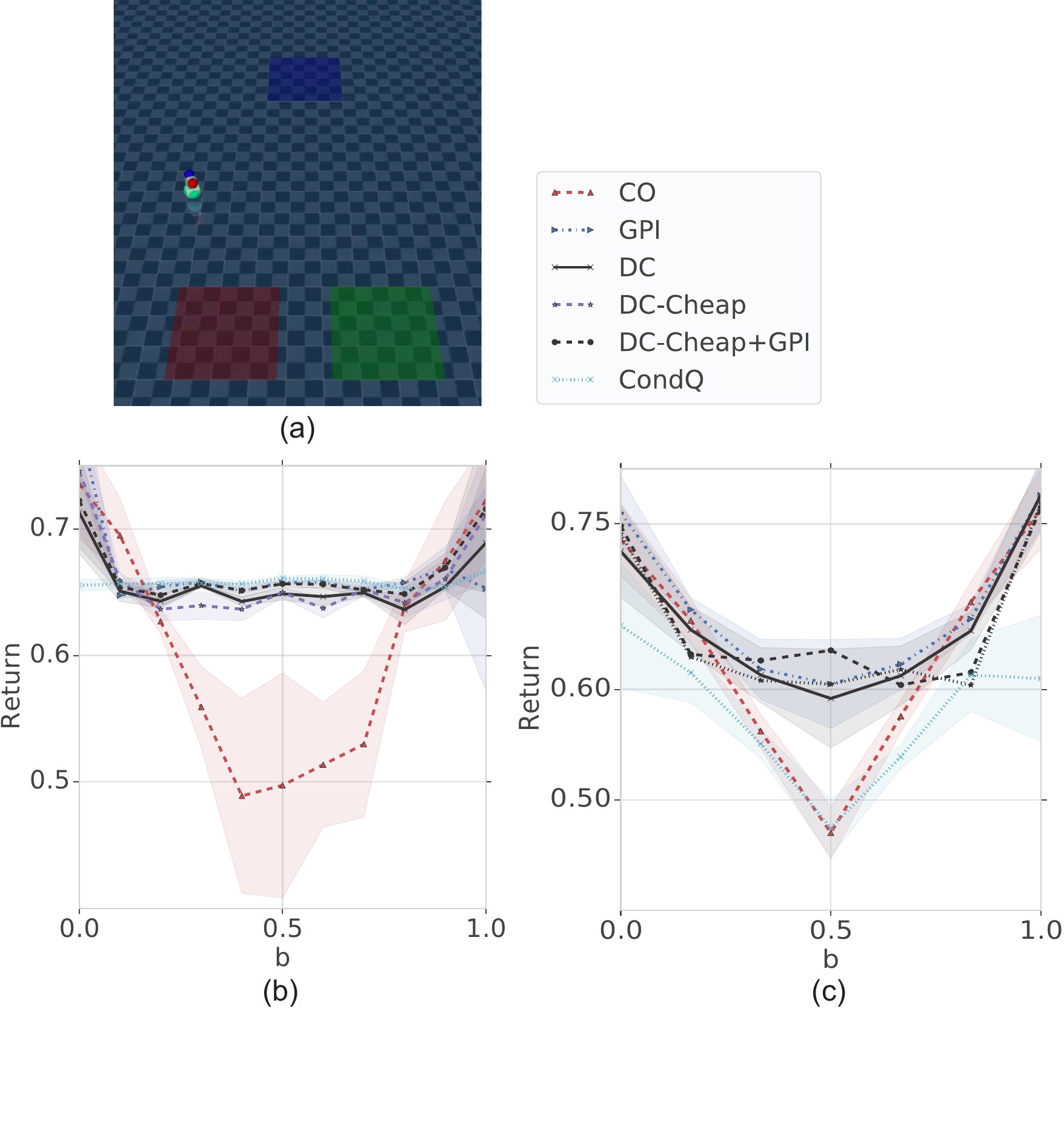}
    \caption{\textbf{Additional results for figure \ref{fig:jumping_ball_tricky} (mobile bodies)}
    \\
    (\textbf{a}) Jumping ball task. The task has rewards $(1, 0), (0, 1)$ in the green and red boxes respectively and $(0.75, 0.75)$ in the blue square.
    \\
    The returns for the transfer task as a function of $b$ ($r_b = b r_1 + (1-b)r_2$) including the DC heuristics for the jumping ball (\textbf{b}) and ant (\textbf{c}). Shaded bars show SEM (5 seeds for ant, 3 seeds for jumping ball).
    As expected, CO performs poorly on these tasks. CondQ struggles to consistently get good returns on the ant task. The DC heuristics perform well on these tasks.
    }
    \label{fig:walker_supp}
\end{figure}

\begin{figure}[H]
    \centering
    \includegraphics[width=\textwidth]{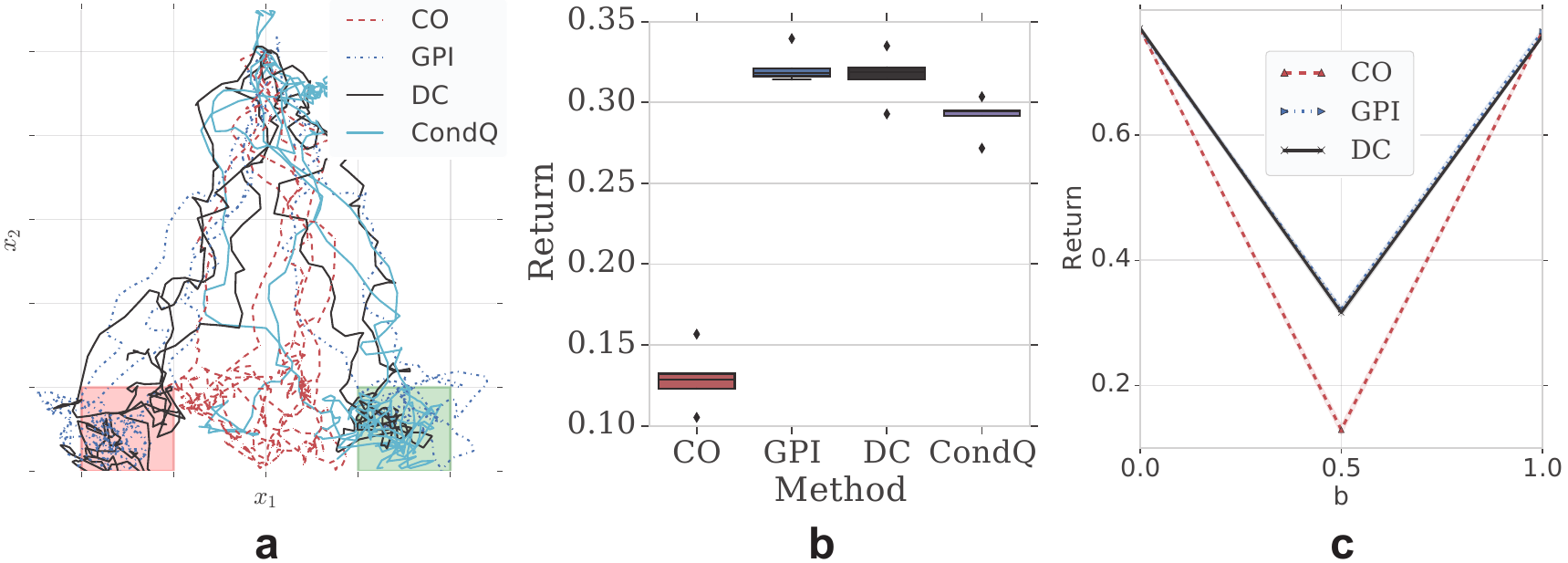}
    \caption{\textbf{Ant on non-composable subtasks}
    \\
    (\textbf{a}) Trajectories of the ant during transfer on non-composable subtasks. In this experiment the two base tasks consists of rewards at the red and green square respectively. As expected, in this task, where the two base tasks have no compositional solution, CO (red) performs poorly with trajectories that end up between the two solutions. GPI (blue) performs well, as does DC (black). CondQ does slightly worse.
    \newline
    (\textbf{b}) Box-plot of returns from 5 seeds (at $b = 0.5$).
    \newline
    (\textbf{c}) Returns as a function of $b$, SEM across 5 seeds is plotted, but is smaller than the line thickness.
    }
    \label{fig:supp_gpi}
\end{figure}

\section{Experiment details} \label{appendix:experiment_details}

All control tasks were simulated using the MuJoCo physics simulator and constructed using the DM control suite \citep{tassa2018deepmind} which uses the MuJoCo physics simulator \citep{todorov2012mujoco}.

The point mass was velocity controlled, all other tasks were torque controlled. The planar manipulator task was based off the planar manipulator in the DM control suite. The reward in all tasks was sparse as described in the main text.

During training for all tasks we start states from the randomly sampled positions and orientations. For the point mass, jumping ball and ant we evaluated transfer starting from the center (in the walker environments, the starting orientation was randomly sampled during transfer, the point mass does not have an orientation). For the planar manipulator transfer was tested from same random distribution as in training. Infinite time horizon policies were used for all tasks.

Transfer is made challenging by the need for good exploration. That was not the focus on this work. We aided exploration in several ways: during training we acted according to a higher-temperature policy $\alpha_{e} = 2\alpha$. We also sampled actions uniformly in an $\epsilon$-greedy fashion with $\epsilon=0.1$ and added Gaussian exploration noise during training. This was sufficient to explore the state space for most tasks. For the planar manipulator and the jumping ball, we found it necessary to induce behavior tasks by learning tasks for reaching the blue target. This behavior policy was, of course, only used for experience and not during transfer.

Below we list the hyper-parameters and networks use for all experiment. The discount $\gamma$ and $\alpha$ were the only sensitive parameters that we needed to vary between tasks to adjust for the differing magnitudes of returns and sensitivity of the action space between bodies. If $\alpha$ is too small then the policies often only find one solution and all transfer approaches behave similarly, while for large $\alpha$ the resulting policies are too stochastic and do not perform well.

\begin{table}[h]
    \centering
    \begin{tabular}{l c}
         Proposal learning rate & $10^{-3}$ \\
         All other learning rates & $10^{-4}$ \\
         Value target update period & $200$ \\
         Proposal target update period & $200$ \\
         $\sfV$ target update period & $500$ \\
         Number of importance samples for all estimators during learning & $200$ \\
         Number of importance samples for acting during training & $50$ \\
         Number of importance samples for acting during transfer & $1000$
    \end{tabular}
    \caption{Parameters the same across all experiments}
\end{table}

The state vector was preprocessed by a linear projection of $3 \times$ its dimension and then a $\tanh$ non-linearity. All action-state networks ($Q$, $\sfQ$, $C$) consisted of 3 hidden layers with $elu$ non-linearities \citep{clevert2015fast}, with both action and preprocessed state projected by linear layers to be of the same dimensionality and used for input the first layer. All value networks and proposal networks consisted of 2 layers with $elu$ non-linearities. The number of neurons in each layer was varied between environments, but was kept the same in all networks and layers  (we did not sweep over this parameter, but choose a reasonable number based on our prior on the complexity of the task).

Below we list the per task hyper-parameters

\begin{table}[h]
    \centering
    \begin{tabular}{l c c c}
         \textbf{Task} & \textbf{Number of units} & \textbf{$\alpha$} & \textbf{$\gamma$}
         \\
         Point mass & 22 & 1 & 0.99 \\
         Planar Manipulator & 192 & 0.05 & 0.99 \\
         Jumping Ball & 192  & 0.2 & 0.9 \\
         Ant & 252 & 0.1 & 0.95 \\
    \end{tabular}
    \caption{Parameters varied between experiments}
\end{table}

\end{document}